\documentclass[runningheads]{llncs}
\usepackage[T1]{fontenc}
\usepackage{csquotes}
\usepackage{graphicx}
\usepackage{subcaption}
\usepackage{hyperref}

\usepackage{mathtools}
\usepackage{amsfonts}
\usepackage{algorithm}
\usepackage{algpseudocodex}
\usepackage{booktabs}
\usepackage{multirow}
\usepackage{array}
\newcolumntype{L}[1]{>{\raggedright\let\newline\\\arraybackslash\hspace{0pt}}m{#1}}
\newcolumntype{C}[1]{>{\centering\let\newline\\\arraybackslash\hspace{0pt}}m{#1}}
\newcolumntype{R}[1]{>{\raggedleft\let\newline\\\arraybackslash\hspace{0pt}}m{#1}}
\usepackage{cite}
\usepackage{layouts}
\usepackage[section]{placeins}

\newcommand{\mat}[1]{\ensuremath{\mathbf{#1}}}
\newcommand{\latentvec}[2]{\ensuremath{\vec{#1}_{#2}}} %
\newcommand{\normaldist}[3]{\ensuremath{\mathcal{N}\mathopen{}\left(#1;#2,#3\right)\mathclose{}}}
\newcommand{\standardnormaldist}{\ensuremath{\mathcal{N}\mathopen{}\left(\vec{0}, \mat{I}\right)\mathclose{}}}
\newcommand{\patchify}[3]{\mbox{\ensuremath{\mathcal{#1}_{p} = \left\{ #2_{p_i}\right\}_{i=1}^{#3} }}}
\newcommand{\append}{\ensuremath{\mathopen{}\mathbin\Vert\mathclose{}}}
\newcommand{\concat}{\ensuremath{\bigoplus}}

\newcommand{\img}[1]{\ensuremath{\prescript{#1}{}{\vec{x}}}} %
\newcommand{\wid}[1]{\ensuremath{\prescript{#1}{}{w}}} %
\newcommand{\writeremb}[1]{\ensuremath{\vec{s}_{#1}}}
\newcommand{\writerembdim}{\ensuremath{d_{s}}}
\newcommand{\timestep}{\ensuremath{t}}
\newcommand{\timestepemb}{\ensuremath{\vec{t}_{emb}}}
\newcommand{\timestepembdim}{\ensuremath{d_{t}}}
\newcommand{\characterembdim}{\ensuremath{d_{e}}}
\newcommand{\contentvec}{\ensuremath{\vec{c}}}
\newcommand{\contentdim}{\ensuremath{d_{c}}}
\newcommand{\contentembseq}[1]{\ensuremath{(\contentvec_1,\dots,\contentvec_{#1})}}
\newcommand{\contentemb}{\ensuremath{\mat{C}}}
\newcommand{\transcr}[1]{\ensuremath{\prescript{#1}{}{\vec{y}}}} %
\newcommand{\transcrseq}[1]{\ensuremath{(y_1, \dots,y_{#1})}} %
\newcommand{\selfattention}[1]{\ensuremath{\text{Self-Attention}\mathopen{}\left( #1 \right)\mathclose{}}}

\newcommand{\vaeenc}[1]{\ensuremath{V_{E}\mathopen{}\left( #1 \right)\mathclose{}}} %
\newcommand{\vaeencname}{\ensuremath{V_{E}}}
\newcommand{\vaedec}[1]{\ensuremath{V_{D}\mathopen{}\left( #1 \right)\mathclose{}}} %
\newcommand{\vaedecname}{\ensuremath{V_{D}}}
\newcommand{\ldm}[1]{\ensuremath{G\mathopen{}\left( #1 \right)\mathclose{}}} %

\newcommand{\mae}[1]{\ensuremath{S_{E}\mathopen{}\left( #1 \right)\mathclose{}}} %
\newcommand{\maename}[1]{\ensuremath{S_{E}}}
\newcommand{\contentenc}[1]{\ensuremath{C_{E}\mathopen{}\left( #1 \right)\mathclose{}}} %
\newcommand{\contentencname}{\ensuremath{C_{E}}}

\newcommand{\contentencts}[1]{\ensuremath{C_{T}\mathopen{}\left( #1 \right)\mathclose{}}} %
\newcommand{\styleenc}[1]{\ensuremath{S_{E}\mathopen{}\left( #1 \right)\mathclose{}}} %
\newcommand{\styleencname}{\ensuremath{S_{E}}}
\newcommand{\maedecname}{\ensuremath{S_{D}}}
\newcommand{\maedec}[1]{\ensuremath{S_{D}\mathopen{}\left( #1 \right)\mathclose{}}} %
\newcommand{\dm}[1]{\ensuremath{\vec{\epsilon}_{\Theta}\mathopen{}\left( #1 \right)\mathclose{}}} %

\DeclarePairedDelimiter{\norm}{\lVert}{\rVert}

\newcommand{\addtotimestep}{\texttt{TS}}
\newcommand{\tokenpreattention}{\texttt{TP}}
\newcommand{\tokenpreattentionlemb}{\texttt{TPL}}
\newcommand{\tokenafterattention}{\texttt{TA}}
\newcommand{\concatpreattentnion}{\texttt{CP}}
\newcommand{\concatafterattention}{\texttt{CA}}

\begin{document}
    \title{Semi-Supervised Adaptation of Diffusion Models for Handwritten Text Generation}
    \subtitle{A Technical Report}
    \titlerunning{Semi-Supervised Adaptation of Diffusion Models for HTG}
    \author{Kai Brandenbusch\orcidID{0000-0001-5258-7454}}
    \authorrunning{K. Brandenbusch}
    \institute{TU Dortmund University, Dortmund, Germany \\
    \email{kai.brandenbusch@tu-dortmund.de}}

    \maketitle %
    \begin{abstract}
The generation of images of realistic looking, readable handwritten text is a challenging task which is referred to as handwritten text generation (HTG).
Given a string and examples from a writer, the goal is to synthesize an image depicting the correctly spelled word in handwriting with the calligraphic style of the desired writer.
An important application of HTG is the generation of training images in order to adapt downstream models for new data sets.
With their success in natural image generation, diffusion models (DMs) have become the state-of-the-art approach in HTG\@.
In this work, we present an extension of a latent DM for HTG to enable generation of writing styles not seen during training by learning style conditioning with a masked auto encoder.
Our proposed content encoder allows for different ways of conditioning the DM on textual and calligraphic features.
Additionally, we employ classifier-free guidance and explore the influence on the quality of the generated training images.
For adapting the model to a new unlabeled data set, we propose a semi-supervised training scheme.
We evaluate our approach on the IAM-database and use the RIMES-database to examine the generation of data not seen during training achieving improvements in this particularly promising application of DMs for HTG\@.
\end{abstract}

    \section{Introduction}
\label{sec:introduction}
The problem of handwritten text generation (HTG) has attracted increasing attention in recent years.
The task is to generate a realistic looking images showing a specified word in handwriting with a desired calligraphic style.
The results can be used to convert digital text into a handwritten document either just for aesthetic reasons or to support people who suffer from writing impairments.
Another promising application is the usage of generated data as training data for handwritten text recognition (HTR) models (e.g.~\cite{Michael2019_ESS,Kang2019_CAS,Li2021_TTb,Retsinas2022_BPH}).
It can either be used to augment the existing training data with more diverse writing styles and an extended vocabulary or be used to help the HTR model adapt to new domains.
One challenge in HTG is that the word on the generated image must be written correctly and legibly which is particularly important when the generated images are used as training material.
An additional goal in HTG is to control the calligraphic style of the generated image or mimic a given writer.

The most simple and scalable approach for HTG is the generation of word images using handwriting like computer fonts (e.g.~\cite{Krishnan2019_Hve}).
However, the computer fonts cannot imitate real handwriting well.
Other approaches extract character glyphs from handwriting samples and use them to synthesize realistic looking word images~\cite{Wang2005_Csp,Lin2007_SpE,Konidaris2007_Kgw,Thomas2009_ShC,Haines2016_MTY} which involves a lot of manual effort.
Therefore, deep learning based methods, mainly Generative Adversarial Networks (GANs)~\cite{Goodfellow2014_GAN,Mirza2014_CGA}, have become increasingly popular in HTG\@.
First proposed in~\cite{Alonso2019_AGH}, systems have been extended by additional loss functions (e.g.~\cite{Davis2020_TSC,Kang2020_GCC,Gan2021_HHI,Wang2022_ABT}), encoders for improved content and style conditioning (e.g.~\cite{Kang2020_GCC,Davis2020_TSC,Krishnan2021_TTT,Pippi2023_HTG}) or new architectures based on transformers~\cite{Vaswani2017_AiA} (e.g.~\cite{Bhunia2021_HT,Wang2022_ABT}).
With the introduction of Denoising Diffusion Probabilistic Models (DDPMs)~\cite{SohlDickstein2015_DUL,Ho2020_DDP}, they became the state of the art in the generation of natural images~\cite{Dhariwal2021_DMB} and achieved impressive results for text-to-image generation~\cite{Nichol2022_GTP,Rombach2022_HRI}.
Some works applied DDPMs for HTG~\cite{Luhman2020_DmH,Zhu2023_CTI,Ding2023_IHO,Nikolaidou2023_WSV,Nikolaidou2024_DTC,Riaz2024_SSA,Dai2024_OSD,Mayr2024_ZSP}.
While DDPMs offer advantages like making loss balancing obsolete and follow a more stable training procedure, the adaptation of DDPMs to HTG poses challenges.
DDPMs are usually generating based on only one conditioning.
For HTG, however, our goal is to condition on the textual content as well as the calligraphic style.
Therefore, writer and text labels are required for the standard training approach of DDPMs and mechanisms to include both conditions have to be developed.
Furthermore, including unlabeled data in the training of DDPMs for HTG, to the best of our knowledge, has not been explored so far.

In this paper, we present a text and style conditional latent diffusion model for semi-supervised adaptation to new datasets.
Following Wordstylist~\cite{Nikolaidou2023_WSV}, we use a variational autoencoder to shift the diffusion process to the latent space.
However, their model can only generate styles of writers seen during training, as they learn style embedding vectors for the writer IDs.
We tackle this limitation by using a masked autoencoder~\cite{He2022_MAA,Souibgui2023_TDs} to extract style embeddings enabling our model to generate handwriting images for unseen writers.
The usage of an MAE for that purpose allows the model to be trained on all available handwriting data without the need for any supervision.
Furthermore, we extend the string encoder of Wordstylist to a content encoder allowing for different ways of incorporating the style conditioning into the text or the timestep conditioning for the diffusion model.
In an extensive evaluation, we measure the performance of the different approaches of including the style for generation of new training images.
For our experiments, we use handwritten text recognition as a downstream task where the model is trained on generated data only.
Using an HTR model for evaluation is a common approach in HTG as it requires the generated images to match the distribution of handwriting styles and, especially, to show correctly spelled text.

The usage of generated data is especially useful for training downstream models in other domains and has been explored for GANs~\cite{Fogel2020_SSS,Gan2021_HHI,Zdenek2021_JME,Bhunia2021_HT,Kang2022_CSA,Zdenek2023_HTG,Chang2023_CHT,Luo2023_SHS,Vanherle2024_VCY} and DDPMs~\cite{Zhu2023_CTI,Mayr2024_ZSP}.
While for GANs semi- or unsupervised training of the generative model has been explored in a few works~\cite{Fogel2020_SSS,Zdenek2021_JME,Zdenek2023_HTG,Chang2023_CHT}, to the best of our knowledge, semi-supervised training including unlabeled data of the target domain has not been applied for DDPMs so far.
We propose a semi-supervised training scheme to extend the generation capabilities of our proposed model to a new dataset.
Besides presenting experiments for generation of a new dataset with a pre-trained model, we show that the semi-supervised training helps to improve generation quality for the new dataset.

In contrast to GANs, DDPMs are usually not trained with an auxiliary classifier for guiding the training.
Instead, classifier-free guidance~\cite{Ho2022_CFD} is often employed~\cite{Ding2023_IHO,Dai2024_OSD,Mayr2024_ZSP}.
During sampling, the guidance scale determines the strength of the influence of the conditioning on the generated image.
For both cases, supervised training for in-domain generation and semi-supervised adaptation to a new dataset, we explore different choices of the guidance scale.
In summary, we make the following contributions:
\begin{itemize}
    \item Extension of~\cite{Nikolaidou2023_WSV} to the generation of unseen writers by conditioning the model on writer embeddings computed by a masked autoencoder
    \item Development of a content encoder with different options for incorporating the style conditioning
    \item Semi-supervised training scheme for adaptation to the generation of new datasets
    \item Evaluation of approaches for including style conditioning and classifier-free guidance for in domain sampling and sampling of unseen datasets
\end{itemize}

The remainder of this paper is structured as follows.
In section~\ref{sec:related-work}, we provide a comprehensive overview of approaches developed for HTG\@.
In section~\ref{sec:dm}, we briefly explain the fundamentals of DDPMs and important extensions to them.
Our proposed system for HTG comprising the latent diffusion model, the MAE for style feature extraction and the content encoder as well as our semi-supervised training scheme is presented in detail in section~\ref{sec:method}.
We describe our evaluation setup and the results of our experiments in section~\ref{sec:experiments} and summarize this work in section~\ref{sec:conclusion}.

    \section{Related Work}
\label{sec:related-work}
In this section, we review the contributions to HTG that have been presented over the years.
We start with an outline of approaches based on extracting and combining character glyphs.
Then we discuss synthesis methods for online handwriting where first methods based on GANs were applied to the problem of HTG\@.
Afterward, we present an overview of methods for offline HTG which were first based on GANs and later on DDPMs.
Finally, we discuss limitations of existing approaches and point out our contributions to tackle these problems.

\subsection{Glyph-based HTG}
\label{subsec:related-work-glyph-based}
First approaches to the challenging task of handwritten text generation (HTG) were based on synthesizing images from a collection of individual character glyphs~\cite{Wang2005_Csp,Lin2007_SpE,Konidaris2007_Kgw,Thomas2009_ShC,Haines2016_MTY}.
Wang et al.~\cite{Wang2005_Csp} learned shape models for each letter from paragraphs of the writer to be imitated and sampled using a conditional delta log-normal model.
For extracting style features, Lin et al.~\cite{Lin2007_SpE} required the user to write isolated characters, special letter pairs and multi-letter words.
Character glyphs are sampled, augmented by geometric deformation, and aligning on a baseline and connected if necessary to render a realistic looking word images.
The synthesis method in~\cite{Konidaris2007_Kgw} requires the user to select and align character templates selected from a document which are then combined to form a synthetic word image.
Thomas et al.~\cite{Thomas2009_ShC} generated CAPTCHAs with synthetic handwriting based on a database of 20.000 character images.
Haines et al.~\cite{Haines2016_MTY} extracted the path of the pen, and segmenting and label it into ligatures and individual glyphs in a semiautomatic fashion followed by an automated synthesis pipeline.
While being able to synthesize realistic looking word images, these methods often require a lot of manual effort for obtaining and preparing the individual character glyphs.
The requirement for expensive manual effort can be avoided when rendering words from true type fonts~\cite{Krishnan2016_GSD,Kang2020_UWA}.
Using 750 publicly available fonts, Krishnan et al.~\cite{Krishnan2016_GSD} created a synthetic dataset containing 9 million images and used it in~\cite{Krishnan2019_Hve} for pre-training a model to learn representations for word spotting.
In their experiments, they showed that there is a domain gap between their synthetic and real word images~\cite{Krishnan2019_Hve} and therefore proposed to use transfer learning to close this gap.
Kang et al.~\cite{Kang2020_UWA} also proposed an approach for writer adaptation from images of rendered words to real handwriting images.

\subsection{Online HTG}
\label{subsec:related-work-online-htg}
In contrast to letter-by-letter HTG, recent approaches generate entire words.
These can be divided into the genration of online and offline handwriting.
In online HTG~\cite{Graves2013_GSR,Aksan2018_DMD,Ganin2018_SPI,Ji2019_GAN,Ingle2019_SHT,Luhman2020_DmH}, the trajectories of the pen are predicted and rendered afterward.
Graves~\cite{Graves2013_GSR} used LSTMs~\cite{Hochreiter1997_LST} conditioned on a text string for generating corresponding pen trajectories.
Writing styles can be imitated by priming the networks memory with a real example of the desired style.
Aksan et al.~\cite{Aksan2018_DMD} proposed to disentangle style and content of handwritten text to enable generation of new words in user-specified styles.
They extended a conditional variational recurrent neural network~\cite{Chung2015_RLV} to generate data from two latent variables to control for content and style.
In contrast, Ganin et al.~\cite{Ganin2018_SPI} proposed an RNN-based GAN~\cite{Goodfellow2014_GAN} for predicting commands for an external rendering engine and demonstrate their approach for the generation of digits and single characters.
Ji et al.~\cite{Ji2019_GAN} improved the LSTM-based model from Graves~\cite{Graves2013_GSR} employing adversarial training with a discriminator.
Another extension of~\cite{Graves2013_GSR} was proposed in~\cite{Ingle2019_SHT}.
An LSTM-based style embedding module is added to allow generation of word images with a desired writing style based on sample images.
A first approach based on DDPMs was proposed by Luhman and Luhman~\cite{Luhman2020_DmH}.
They conditioned the diffusion model on a sequence of character embeddings and a feature sequence computed by cross-attention between the character embeddings and style features extracted by a MobileNetV2~\cite{Sandler2018_MIR}.

\subsection{Offline HTG}
\label{subsec:related-work-offline-htg}
Despite the promising results achieved by online HTG, the approaches suffer from a fundamental problem.
Temporal data recorded by a digital stylus pen is required for training which is costly to obtain.
Therefore, the majority of methods for HTG rely on (offline) images of handwritten text.

\subsubsection{GAN-based}
\label{subsubsec:related-work-offline-gan}
The development in this area was initially driven by the success GANs~\cite{Goodfellow2014_GAN,Mirza2014_CGA} in image generation.
Alonso et al.~\cite{Alonso2019_AGH} used a GAN conditioned on string encodings obtained by an LSTM~\cite{Hochreiter1997_LST}.
Besides the adversarial objective, an auxiliary handwritten text recognition (HTR) model with CTC-Loss~\cite{Graves2006_Ctc} is employed for training in order to enforce correctness of the generated text.
ScrabbleGAN~\cite{Fogel2020_SSS} uses a similar adversarial setup with an HTR model.
Instead of a global word representation, the model uses one filter for each character for the textual conditioning.
Style variation is introduced by multiplication of the character filter with a random noise vector.
However, the style cannot be controlled in both works~\cite{Alonso2019_AGH,Fogel2020_SSS}.
To enable the generation of handwriting in a user-specified style, GANwriting~\cite{Kang2020_GCC} computes style embeddings with a style encoder from 15 example images.
To force the generator to learn the correct generation for the styles, an auxiliary writer classifier is used for training.
For the textual conditioning, a character-wise as well as a global embedding is computed for a given string.
In another work~\cite{Kang2020_DCS}, Kang et al.\ also proposed a GAN where the textual conditioning is given by features extracted by a recognition model from an example image containing the desired word.
Later, the approach~\cite{Kang2020_GCC} was extended~\cite{Kang2022_CSA} to generate whole lines instead of words.
Davis et al.~\cite{Davis2020_TSC} presented a GAN which is able to generate words as well as lines of text.
With their style extractor acting as an encoder, they trained the model together with the generator like an autoencoder using a perceptual and a reconstruction loss while also employing an adversarial objective and a CTC-loss to obtain legible handwriting.
Additionally, a spacing model is used for better alignment of the generated image.

A different approach for the textual conditioning was employed by Guan et al.~\cite{Guan2020_IHO} who used skeleton images extracted by an external tool.
Style information is inferred from another example image with the desired style.
In contrast to previous methods, the generated content is not checked by an HTR model.
Instead, an L1-loss between a generated and a real image with the same style and content is computed.
Mayr et al.~\cite{Mayr2020_STH} proposed a similar approach of generation from an extracted skeleton.
Online data is obtained by skeletonization followed by an estimation of temporal information.
The first style transfer is carried out applying~\cite{Graves2013_GSR} to the online data.
After rendering the generated strokes, the image style (e.g.\ ink) is transferred using~\cite{Isola2017_IIT}.

Like ScrabbleGAN~\cite{Fogel2020_SSS}, HiGAN~\cite{Gan2021_HHI} starts generation based on a text-map to enable the generation of long texts.
By conditioning the generator on features computed by a style encoder, it is possible to control the style.
A KL-Divergence regularization of the style vector allows to sample random styles from a normal distribution.
An HTR model and a writer identifier are used to guide the training.
JokerGAN~\cite{Zdenek2021_JME} is also inspired by ScrabbleGAN~\cite{Fogel2020_SSS}.
The filter bank of the generator is replaced by a base filter for all characters and the conditioning is incorporated by the proposed multi-class conditional batch normalization which is an extension of~\cite{Miyato2018_cPD}.
Additionally, the generator is conditioned on vertical layout information based on baseline, mean line, ascenders and descenders in the target word.
The model was later improved by a more sophisticated discriminator and a style encoder to control the style of the generated image~\cite{Zdenek2023_HTG}.
Mattick et al.~\cite{Mattick2021_SIH} extended GANwriting~\cite{Kang2020_GCC} by an additional local discriminator in order to reduce pen-level artifacts in the generated images.
Liu et al.~\cite{Liu2021_HTG} trained their GAN to reconstruct a word image based on a text conditioning and a style embedding extracted from an example image.
At the same time, the model is trained to generate another word with the same style.
Thereby, the authors could enhance the visual quality by employing an adversarial as well as a reconstruction loss without the need for writer labels.
TextStyleBrush~\cite{Krishnan2021_TTT} also does only require textual labels for training by exploiting cycle consistency and reconstruction losses for guiding the model to learn generation in the desired style.

Instead of using convolutional generator like most other approaches at that time, HWT~\cite{Bhunia2021_HT} is based on transformers~\cite{Vaswani2017_AiA}.
An encoder with self-attention computes a style feature sequence which is used as key-value pairs in the cross-attention of the decoder with character embeddings of the desired string as queries.

By adding a local patch loss, a contextual style loss and reusing features from the writer identifier as style features, HiGAN++~\cite{Gan2022_HHI} was presented as an improvement of HiGAN~\cite{Gan2021_HHI}.
Wang et al.~\cite{Wang2022_ABT} improved GANwriting~\cite{Kang2020_GCC} by replacing the generator with a transformer-based model.
Additionally, they proposed a feature deformation fusion module for an improved local style extraction and the use of focal frequency loss to solve stroke level artifact problems.

While other works focus on generating handwriting in the language the model was trained on, Chang et al.~\cite{Chang2023_CHT} employed an image-to-image approach to generate images of words from an unseen language.
SLOGAN~\cite{Luo2023_SHS} uses a style bank to encode styles from writers in the training set.
They employ two discriminators.
The local discriminator is focusing on the generation quality of individual characters localized by attention an auxiliary HTR head using a recognition and an adversarial loss.
Another global discriminator is used to control the style of the generated word by an adversarial and a writer identification loss.
Additionally, the generator is regularized by an autoencoder constraint (L2 loss) for generated and real images with the same content and style.
The visual archetypes-bases transformer (VATr)~\cite{Pippi2023_HTG} is a transformer-based model with a similar idea as~\cite{Bhunia2021_HT}.
A sequence of style vectors is extracted from example images by a pre-trained convolutional feature extractor followed by a transformer with self-attention.
In contrast to~\cite{Bhunia2021_HT}, for content conditioning the words are rendered to binary images using GNU Unifont.
After flattening and projecting these images, a transformer decoder performs cross-attention between the content and the style sequences.
A convolutional decoder generated the final image based on the resulting representation.
In addition to a discriminator, and HTR model and a style classifier are used to guide the training.
Vanherle et al.~\cite{Vanherle2024_VCY} proposed small improvements to~\cite{Pippi2023_HTG} by a better handling of punctuation marks, augmentations to balance the character distribution during training and improved regularization by the discriminator and the HTR model.

\subsubsection{DDPM-based}
\label{subsubsec:related-work-offline-ddpm}
With achieving better generation quality than GANs for the generation of natural images~\cite{Dhariwal2021_DMB}, denoising diffusion probabilistic models~\cite{SohlDickstein2015_DUL,Ho2020_DDP} (DDPM) have become more popular for HTG\@.
A method for offline HTG with DDPMs was proposed by Zhu et al.~\cite{Zhu2023_CTI}.
The DDPM can be conditioned on different combinations of encodings of the text, the style and the image.
The image conditioning is obtained by attention pooling~\cite{Lee2019_STF} of features from a pre-trained text recognizer, therefore encoding the unique visual characteristics of the image.
The text condition is obtained by multiplying the classifier weights of the pre-trained text recognizer with the one-hot encoded characters of a string.
The personal style of writers is encoded by a learned embedding for each writer.
Depending on which conditions are given, the model can be used for synthesis, augmentation, recovery and imitation.
Based on a model for generation of handwritten chinese character~\cite{Gui2023_ZsG}, Ding et al.~\cite{Ding2023_IHO} conditioned their DDPM on a glyph image and the writer ID\@.
In order to reduce the computational costs of DDPMs, latent diffusion models~\cite{Rombach2022_HRI} (LDMs) were proposed which perform the diffusion process in a latent space.
Wordstylist~\cite{Nikolaidou2023_WSV} is based on an LDM which is conditioned on an encoded string and learned embeddings for the writers from the training set.
The latent representation at the end of the reverse diffusion process is then used by a pre-trained VAE~\cite{Rombach2022_HRI} to generate the final handwriting image.
DiffusionPen~\cite{Nikolaidou2024_DTC} improves upon Wordstylist~\cite{Nikolaidou2023_WSV} by using an advanced text encoding and a style encoder.
The style encoder is a MobileNetV2~\cite{Sandler2018_MIR} pre-trained with a classification and a triplet loss for learning a style representation.
Riaz et al.~\cite{Riaz2024_SSA} combined the ideas of computing a text embedding with character level tokenization and a transformer encoder (e.g.~\cite{Nikolaidou2023_WSV}) and conditioning on a rendered glyph image (e.g.~\cite{Ding2023_IHO}) of the word to be generated.
For conditioning on the style, however, they only learn embeddings for the writers seen during training.
Dai et al.~\cite{Dai2024_OSD} argued that the generation quality can be improved by giving more focus to high-frequency information from the style example images and propose a style-enhanced module comprising two style encoders.
One encoder extracts features from the high-frequency components computed by a Laplacian kernel the style image.
The style features from the other encoder computed on the original style image are passed through a gating mechanism to filter out background noise.
The text embedding is computed by a transformer from character-wise ResNet18 features from an image rendered in unifont.
The final conditioning is computed by a style-content fusion module using cross-attention.
In contrast to previous works, Mayr et al.~\cite{Mayr2024_ZSP} are the first to generate whole paragraphs using LDMs.
Instead of using a VAE pre-trained on natural images~\cite{Rombach2022_HRI}, they retrained the VAE on few real samples using additional losses for HTR and writer identification to enforce a stronger compression of document-relevant information.
Style and text are encoded separately and then fused using cross-attention.

\subsection{Discussion}
\label{subsec:related-work-discussion}
The training of GANs has been heavily influenced by the choice of auxiliary models and their loss functions.
Besides the adversarial loss, mostly an HTR model was used to enforce correct content~\cite{Alonso2019_AGH,Fogel2020_SSS,Kang2020_GCC,Kang2020_DCS,Davis2020_TSC,Gan2021_HHI,Zdenek2021_JME,Bhunia2021_HT,Mattick2021_SIH,Liu2021_HTG,Krishnan2021_TTT,Kang2022_CSA,Gan2022_HHI,Wang2022_ABT,Zdenek2023_HTG,Chang2023_CHT,Luo2023_SHS,Pippi2023_HTG,Vanherle2024_VCY}
while a writer classifier was used to force generation in the desired style~\cite{Kang2020_GCC,Kang2020_DCS,Gan2021_HHI,Bhunia2021_HT,Mattick2021_SIH,Kang2022_CSA,Gan2022_HHI,Wang2022_ABT,Zdenek2023_HTG,Luo2023_SHS,Pippi2023_HTG,Vanherle2024_VCY}.
For additional regularization, some works also employed reconstruction losses\cite{Davis2020_TSC,Guan2020_IHO,Mayr2020_STH,Liu2021_HTG,Krishnan2021_TTT,Zdenek2023_HTG,Luo2023_SHS} or cycle consistency of images or extracted embeddings\cite{Gan2021_HHI,Bhunia2021_HT,Krishnan2021_TTT,Gan2022_HHI,Pippi2023_HTG,Vanherle2024_VCY}.
However, since the training of DDPMs involves the generation of images at different noise levels, using an auxiliary model is not suitable.
Therefore, classifier-free guidance was proposed~\cite{Ho2022_CFD} and adopted for some approaches~\cite{Ding2023_IHO,Dai2024_OSD,Mayr2024_ZSP} using DDPMs for HTG\@.
In this work, we explore the influence of guided sampling for an LDM trained using classifier-free guidance.

Many DDPM-based approaches~\cite{Zhu2023_CTI,Ding2023_IHO,Nikolaidou2023_WSV,Riaz2024_SSA} for HTG condition their model on a learned embedding vector per writer which is mapped from its writer ID\@.
As a consequence, these models are only able to generate styles from writers seen during training or random styles.
To specifically imitate the style of a new, previously unseen writer, other works~\cite{Luhman2020_DmH,Nikolaidou2024_DTC,Dai2024_OSD,Mayr2024_ZSP} employ a style encoder for computing style embeddings based on example images.
These models are trained on samples labeled with writer IDs except for~\cite{Luhman2020_DmH} who use a model which was pre-trained on natural images.
While requiring no labeled handwriting examples, it is doubtful how well the model extracts handwriting specific features.
Similar to DiffusionPen~\cite{Nikolaidou2024_DTC} which was developed independently at the same time as this work, we extend~\cite{Nikolaidou2023_WSV} with a style encoder to enable generation of new styles.
\cite{Nikolaidou2024_DTC} trained their style encoder with a classification and a triplet loss forcing the model to focus on calligraphic features.
In contrast, we utilize a masked autoencoder~\cite{He2022_MAA,Souibgui2023_TDs} for computing style embeddings.
Due to the reconstruction task, we expect our model to encode both, textual and calligraphic content, resulting in an inferior style embedding.
However, the major advantage of using an MAE is that it allows us to include arbitrary handwriting examples without being restricted to labeled examples like~\cite{Nikolaidou2024_DTC}.
To further leverage the capability of generating unseen writing styles, we propose a semi-supervised training scheme of our DDPM to improve generation of examples from new datasets.

    \section{Diffusion Models}
\label{sec:dm}
Denoising diffusion probabilistic models (DDPMs)~\cite{SohlDickstein2015_DUL,Ho2020_DDP} are a class of generative models yielding state-of-the-art results in image generation~\cite{Dhariwal2021_DMB}.
The diffusion process is a Markov chain where noise is gradually added to the input data~\cite{SohlDickstein2015_DUL}.
Thereby, a complex data distribution $q(\latentvec{x}{0})$ is converted into an analytically tractable distribution e.g.\ a standard normal distribution.
Samples are then generated by reversing this process with learned transitions~\cite{Ho2020_DDP}.
In this section, we review the forward and reverse diffusion process, the training of DDPMs and extensions to conditional and latent DDPMs.

\subsection{Forward and Reverse Process}
\label{subsec:dm-forward-and-reverse}
The addition of noise is done in the so-called forward process (or diffusion process).
It is defined as a Markov chain where at each timestep $t=1,\dots,T$ Gaussian noise is added to the data $\latentvec{x}{0}  \sim q\left( \latentvec{x}{0} \right)$ with a defined variance schedule $\beta_t \in \left( 0,1 \right)$~\cite{Ho2020_DDP}:
\begin{align}
    \label{eq:dm-forward-joined}
    q\left(\latentvec{x}{1},\dots,\latentvec{x}{T} |\latentvec{x}{0}\right) & \coloneqq \prod_{t=1}^T q\left(\latentvec{x}{t} | \latentvec{x}{t-1}\right) \\
    \label{eq:dm-forward-step}
    q\left(\latentvec{x}{t} | \latentvec{x}{t-1}\right) & \coloneqq \normaldist{\latentvec{x}{t}}{\sqrt{1-\beta}\latentvec{x}{t-1}}{\beta_t\mathbf{I}}
\end{align}
where $\latentvec{x}{1},\dots,\latentvec{x}{T}$ are latent variables.
For a sufficiently large $T$ and a suitable variance schedule $\beta_t \in (0, 1)$, the latent $\latentvec{x}{T}$ follows almost a standard Gaussian distribution $\standardnormaldist$~\cite{Nichol2021_IDD}.

For sampling, the process has to be reversed by a stepwise removal of noise.
The process would require $q\left(\latentvec{x}{t-1} | \latentvec{x}{t}\right)$, which depends on the entire data distribution~\cite{Nichol2021_IDD}.
Therefore, a model $p_{\Theta}\left( \latentvec{x}{t-1} | \latentvec{x}{t} \right)$ is trained to approximate this distribution.
Starting at $\latentvec{x}{T} \sim \standardnormaldist$, the transitions of the reverse process are defined as:
\begin{align}
    \label{eq:dm-reverse-joined}
    p_{\Theta}\left(\latentvec{x}{0},\dots,\latentvec{x}{T}\right) & \coloneqq p(\latentvec{x}{T}) \prod_{t=1}^T p_{\Theta}\left( \latentvec{x}{t-1} | \latentvec{x}{t} \right) \\
    \label{eq:dm-reverse-step}
    p_{\Theta}\left( \latentvec{x}{t-1} | \latentvec{x}{t} \right) & \coloneqq \normaldist{\latentvec{x}{t-1}}{\vec{\mu}_\Theta\left( \latentvec{x}{t},t \right)}{\mat{\Sigma}_\Theta\left( \latentvec{x}{t},t \right)}.
\end{align}

\subsection{Training}
\label{subsec:dm-training}
The model is trained by optimizing the variational bound of the negative log-likelihood of the training data~\cite{Ho2020_DDP}:
\begin{equation}
    \label{eq:dm-training-nll}
    \mathbb{E}\left[ -\log\left(p_{\Theta}\left(\latentvec{x}0{}\right)\right) \right] \le \mathbb{E}_q \left[ -\log \frac{p_{\Theta}\left(\latentvec{x}{0},\dots,\latentvec{x}{T}\right)}{q\left(\latentvec{x}{1},\dots, \latentvec{x}{T} | \latentvec{x}{0}\right)} \right] \eqqcolon L.
\end{equation}
The loss can be reformulated as
\begin{equation}
    \label{eq:dm-training-nll-reformulated}
    L = \mathbb{E}_q \left[ -\log p\left( \latentvec{x}{T}\right) -\sum_{t=1}^{T} \log \frac{p_{\Theta}\left(\latentvec{x}{t-1} | \latentvec{x}{t}\right)}{q\left(\latentvec{x}{t} | \latentvec{x}{t-1} \right)} \right].
\end{equation}
After further reformulations and conditioning the forward process $q\left(\latentvec{x}{t} | \latentvec{x}{t-1} \right)$ on the initial sample $\latentvec{x}{0}$ (see~\cite{Ho2020_DDP} for details), the following objective is obtained
\begin{equation}
    \label{eq:dm-training-nll-conditionied-reformulated}
    \begin{aligned}
        L = \mathbb{E}_q \Biggl[ & D_{KL}\left( q\left(\latentvec{x}{T} | \latentvec{x}{0}\right) || p\left( \latentvec{x}{T} \right) \right) \\
        & + \sum_{t=2}^T D_{KL}\left( q\left(\latentvec{x}{t-1} | \latentvec{x}{t}, \latentvec{x}{0}\right) || p_{\Theta}\left( \latentvec{x}{t-1} | \latentvec{x}{t} \right) \right) - \log p_{\Theta}\left( \latentvec{x}{0} | \latentvec{x}{1} \right) \Biggr],
    \end{aligned}
\end{equation}
where $D_{KL}$ denotes the KL divergence.
The definition of the forward process in Eq.~\ref{eq:dm-forward-step} allows for sampling $\latentvec{x}{t}$ at an arbitrary timestep $t$ in closed form by
\begin{equation}
    \label{eq:dm-training-closed-form-sampling}
    q\left(\latentvec{x}{t} | \latentvec{x}{0}\right) = \normaldist{\latentvec{x}{t}}{\sqrt{\bar{\alpha}_t}\latentvec{x}{0}}{\left( 1-\bar{\alpha}_t \right)\mat{I}}
\end{equation}
where $\alpha_t \coloneqq 1-\beta_t$ and $\bar{\alpha}_t \coloneqq \prod_{s=1}^t \alpha_s$.
Therefore, $q(\latentvec{x}{t-1} | \latentvec{x}{t}, \latentvec{x}{0})$ can be computed as
\begin{align}
    q(\latentvec{x}{t-1} | \latentvec{x}{t}, \latentvec{x}{0}) &= \normaldist{\latentvec{x}{t-1}}{\tilde{\vec{\mu}}_t\left( \latentvec{x}{t}, \latentvec{x}{0} \right)}{\tilde{\beta}_t\mat{I}} \\
    \text{ with } \tilde{\vec{\mu}}_t\left( \latentvec{x}{t}, \latentvec{x}{0} \right) &\coloneqq \frac{\sqrt{\bar{\alpha}_{t-1}}\beta_t}{1-\bar{\alpha}_t}\latentvec{x}{0}
    + \frac{\sqrt{\alpha_t}\left( 1-\bar{\alpha}_{t-1} \right)}{1-\bar{\alpha}_t}\latentvec{x}{t} \text{ and } \tilde{\beta}_t \coloneqq \frac{1 - \bar{\alpha}_{t-1}}{1 - \bar{\alpha}_t} \beta_t.
\end{align}
Setting the variance to a fixed schedule in the forward process, $p_{\Theta}\left( \latentvec{x}{t-1}|\latentvec{x}{t} \right)$ can be parametrized by
\begin{equation}
    \label{eq:dm-forward-step-fixed-variance}
    p_{\Theta}\left( \latentvec{x}{t-1}|\latentvec{x}{t} \right) = \normaldist{\latentvec{x}{t-1}}{\vec{\mu}_{\Theta}\left( \latentvec{x}{t}, t \right)}{\beta_t\mat{I}}.
\end{equation}
From $D_{KL}\left( q(\latentvec{x}{t-1} | \latentvec{x}{t}, \latentvec{x}{0}) || p_{\Theta}\left( \latentvec{x}{t-1} | \latentvec{x}{t} \right) \right)$ it can be derived (see~\cite{Ho2020_DDP} for details) that
the model must then be trained to predict the mean $\vec{\mu}_{\Theta}\left( \latentvec{x}{t}, t \right)$ with
\begin{equation}
    \label{eq:mu_pred}
    \vec{\mu}_{\Theta}\left( \latentvec{x}{t}, t \right) = \frac{1}{\sqrt{\alpha_t}}\left( \latentvec{x}{t} - \frac{\beta_t}{\sqrt{1-\bar{\alpha}_t}} \vec{\epsilon}_{\Theta}\left( \latentvec{x}{t}, t \right) \right).
\end{equation}
Equivalently, the model can directly predict the noise $\vec{\epsilon}_{\Theta}(\latentvec{x}{t}, t)$ added to the image at timestep $t$.
This parametrization resembles Langevin dynamics and denoising score matching over multiple noise scales~\cite{Ho2020_DDP,Song2019_GME}.
Omitting scaling terms that have been found to be not beneficial for sampling quality~\cite{Ho2020_DDP}, the model can be trained using the following simplified objective:
\begin{equation}
    \label{eq:dm-training-simple-loss}
    L_{\text{simple}}\left( \Theta \right) \coloneqq \mathbb{E}_{t, \latentvec{x}{0}, \epsilon} \left[ \norm{\vec{\epsilon} - \vec{\epsilon}_{\Theta}\left( \latentvec{x}{t}, t \right)}^2 \right].
\end{equation}
The objective is then optimized by stochastic gradient descent on uniformly sampled timesteps $t$.

\subsection{Sampling}
\label{subsec:dm-sampling}
For the generation of images, $\latentvec{x}{t-1} \sim p_{\Theta}\left( \latentvec{x}{t-1}|\latentvec{x}{t} \right)$ has to be sampled~\cite{Ho2020_DDP}.
Given a random latent $\vec{n} \sim \mathcal{N}\left( \vec{0}, \mat{I} \right)$ and the noise prediction $\vec{\epsilon}_{\Theta}\left( \latentvec{x}{t}, t \right)$ from the network,
$\latentvec{x}{t-1}$ can be computed from $\latentvec{x}{t}$ by
\begin{equation}
    \label{eq:dm-sampling-basic}
    \latentvec{x}{t-1} = \frac{1}{\sqrt{\alpha}_t} \left( \latentvec{x}{t} - \frac{\beta_t}{\sqrt{1 - \bar{\alpha}}}_t \vec{\epsilon}_{\Theta}\left( \latentvec{x}{t}, t \right) \right) + \beta_t\vec{n}.
\end{equation}
This removal of noise is repeated, starting at timestep $t=T$, in order to obtain a generated image $\latentvec{x}{0}$ following algorithm~\ref{alg:sampling}.
\begin{algorithm}
    \caption{Sampling~\cite{Ho2020_DDP}}\label{alg:sampling}
    \begin{algorithmic}
        \State $\latentvec{x}{T} \sim \mathcal{N}\left( \vec{0}, \mat{I} \right)$
        \For{$t=T,\dots,1$}
            \State $\vec{n} \sim \mathcal{N}\left( \vec{0}, \mat{I} \right)$ if $t > 1$, else $\vec{n}=\vec{0}$
            \State $\latentvec{x}{t-1} = \frac{1}{\sqrt{\alpha}_t} \left( \latentvec{x}{t} - \frac{\beta_t}{\sqrt{1 - \bar{\alpha}}}_t \vec{\epsilon}_{\Theta}\left( \latentvec{x}{t}, t \right) \right) + \beta_t\vec{n}$
        \EndFor
        \Return $\latentvec{x}{0}$
    \end{algorithmic}
\end{algorithm}

This sampling strategy requires the simulation of a Markov chain for many (usually $T=1000$) steps~\cite{Song2021_DDI} making the sampling process computationally very expensive.
Therefore, new approaches for sampling have been researched.
Song et al.~\cite{Song2021_DDI} propose a generalization of DDPMs by a class of non-Markovian diffusion processes.
While the training objective remains the same, their approach allows to consider forward processes with less than $T$ steps resulting in a reverse process with correspondingly fewer steps.
Thereby, DDPMs trained as described in the previous section can be used and sampled in a more efficient way.
Recent approaches further reduced the number of sampling steps required to generate high quality images~\cite{Lu2022_DSF,Lu2022_DSFa,Zhang2023_FSD,Zhao2023_UUP}.

\subsection{Conditional Generation}
\label{subsec:dm-conditional-sampling}
The approach described so far only allows for unconditional generation.
To be able to utilize the full potential of generative models, control over the content of the generation is required.
This is realized by conditioning the model on class labels~\cite{Dhariwal2021_DMB,Ho2022_CFD,Song2021_SBG}, text prompts~\cite{Nichol2022_GTP} or other modalities.
Sohl-Dickstein et al.~\cite{SohlDickstein2015_DUL} proposed to modify the model distribution $p_{\Theta}$ in the reverse process by multiplication with another distribution.
Following the idea of Song et al.~\cite{Song2021_SBG} for conditional sampling for score based generative models, Dhariwal et al.~\cite{Dhariwal2021_DMB} proposed to use a classifier for guiding the sampling process of DDPMs.
First, a classifier $c_{\Phi}\left( y | \latentvec{x}{t}, t \right)$ is trained on noise images $\latentvec{x}{t}$ where $y$ is the class label.
Then, $\vec{\epsilon}_{\Theta}\left( \latentvec{x}{t}, t \right)$ can be modified by addition of the gradient of the classifier
\begin{equation}
    \label{eq:dm-conditional-sampling-noise-class-grad}
    \vec{\hat{\epsilon}}_{\Theta}\left( \latentvec{x}{t}, t \right) = \vec{\epsilon}_{\Theta}\left( \latentvec{x}{t}, t \right) - \sqrt{1 - \bar{\alpha}_t} w_{gs} \nabla_{\latentvec{x}{t}}\log c_{\Phi}\left( y | \latentvec{x}{t}, t \right)
\end{equation}
where $w_{gs}$ is a weight to scale the influence of the guidance.
Thereby, sampling is guided toward the classifier predicting the given conditioning $y$.
Dhariwal et al.\ found that conditioning the diffusion model on the class label in addition to guiding the sampling further improves generation quality.

However, classifier guidance requires an additional model to be trained since pre-trained classifiers are usually not available for noisy data.
Classifier-free (diffusion) guidance~\cite{Ho2022_CFD} is an approach avoiding the use of a classifier.
Instead, besides the unconditional diffusion model parametrized by $\vec{\epsilon}_{\Theta}\left( \latentvec{x}{t}, t \right)$ a conditional model parametrized by $\vec{\epsilon}_{\Theta}\left( \latentvec{x}{t}, t, y \right)$ is trained.
Ho et al.\ proposed to use a single model where the conditioning is set to a null label $\emptyset$ for the unconditional case:
\begin{equation}
    \label{eq:dm-cfg-epsilon-pred-uncond}
    \vec{\epsilon}_{\Theta}\left( \latentvec{x}{t}, t \right) = \vec{\epsilon}_{\Theta}\left( \latentvec{x}{t}, t, y=\emptyset \right).
\end{equation}
During training, the conditioning is set to the null label with some probability $p_{\text{uncond}}$.
For sampling, a linear combination of conditional and unconditional predictions is used:
\begin{equation}
    \label{eq:dm-cfg-epsilon-pred-guidance}
    \vec{\hat{\epsilon}}_{\Theta}\left( \latentvec{x}{t}, t, y \right) = (1 + w_{gs})\vec{\epsilon}_{\Theta}\left( \latentvec{x}{t}, t, y \right) - w_{gs}\vec{\epsilon}_{\Theta}\left( \latentvec{x}{t}, t, y=\emptyset \right)
\end{equation}
where $w_{gs}$ determines the strength of the guidance.

\subsection{Latent Diffusion Models}
\label{subsec:dm-latent-dm}
While achieving state-of-the-art generation performance, training and inference of diffusion models is extremely expensive regarding GPU compute power and time.
Rombach et al.\cite{Rombach2022_HRI} proposed, to separate training into two stages.
First, an autoencoder based on~\cite{Esser2021_TTH} is trained to provide a lower-dimensional latent space.
Given an RGB image $\vec{x} \in \mathbb{R}^{H\times W\times 3}$, the encoder $\vaeencname$ computes the latent space representation $\vec{z} \in \mathbb{R}^{h \times w \times c}$ with $f=H/h = W/w$ being the downsampling factor.
A decoder $\vaedecname$ reconstructs the image $\tilde{\vec{x}} = \vaedec{\vec{z}} = \vaedec{\vaeenc{\vec{x}}}$ from latent representation.
The model is trained using a perceptual loss~\cite{Zhang2018_UED} and a path-based adversarial loss~\cite{Isola2017_IIT}.
Additionally, Rombach et al.\ tried a KL-penalty and a vector quantization layer~\cite{Oord2017_NDR} as regularization.
Then, instead of training the diffusion model in the high-dimensional RGB image space $\mathbb{R}^{H\times W\times 3}$, the model is trained in the low-dimensional latent space $\mathbb{R}^{h \times w \times c}$~\cite{Rombach2022_HRI}.
Therefore, these models are denoted as latent diffusion models (LDMs).
Another advantage of the two-stage training is that the trained autoencoder can be reused for many different generative applications.

    \section{Method}
\label{sec:method}
Our proposed approach works on a multi-writer dataset $\left\{ \mathcal{X}, \mathcal{W}, \mathcal{Y} \right\}$ with $N$ images $\mathcal{X}=\left\{ \img{i} \right\}_{i=1}^N$.
Every image $\img{i}$ is labeled with its corresponding writer identifier $\wid{i}$ and its transcription $\transcr{i}=\transcrseq{l_i}$.
We use left superscript for indexing the dataset to avoid confusion with the timestep index in the diffusion process.
Further, we omit the timestep index $t$ if $t=0$, e.g.\ $\img{i} = \prescript{i}{}{\latentvec{x}{0}}$ for better legibility when not describing the diffusion process itself.
Given some writer $\wid{} \in \mathcal{W}$, we refer to the subset of all images from $\wid{}$ as $\mathcal{X}_{\wid{}} = \left\{ \img{i} | \wid{i}=\wid{} \right\} \subset \mathcal{X}$ with $N_{\wid{}} = |\mathcal{X}_{\wid{}}|$ and
$\mathcal{X}_{\wid{}}^{K} \subset \mathcal{X}_{\wid{}}$ being a subset of $K \le N_{\wid{}}$ random examples from writer $\wid{}$.
We define $\mathcal{A}^l$ to be the set of all possible text strings of length $l$ made up of a selection of allowed characters (e.g.\ letters, digits, punctuation).
The task is to generate an image $\hat{\vec{x}}=\ldm{\transcr{}, \mathcal{X}_{\wid{}}^{K}}$ given a transcription $\transcr{} \in \mathcal{A}^l$ and a set of example images $\mathcal{X}_{\wid{}}^{K}$ from some writer $\wid{}$.
The generated image should depict the handwritten word with transcription $\transcr{}$ while resembling the calligraphic style of writer $w$.
Note that $\transcr{}$ may not be in $\mathcal{Y}$ (\textit{out-of-vocabulary}) and $\wid{}$ might not be in $\mathcal{W}$ (\textit{unseen writer}).

\subsection{Overview}
\label{subsec:method-overview}
\begin{figure}[tb]
    \centering
    \includegraphics[width=\linewidth]{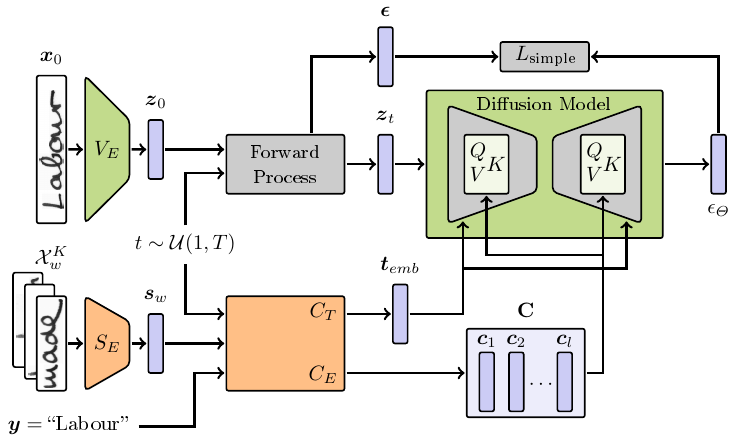}
    \caption{
        Overview of the proposed system during training given an input image $\latentvec{x}{0}$, its transcription $\transcr{}$ and a set $\mathcal{X}_{w}^{K}$ of examples from the sample writer.
        The system comprises the encoder part $\vaeencname$ of the VAE, style encoder $\styleencname$, content encoder $\contentencname$ and the LDM.
        The LDM $\dm{\latentvec{z}{t}, \timestepemb, \contentemb}$ is trained to predict the noise added in step $\timestep$ of the forward diffusion process.}
    \label{fig:method-overview-train}
\end{figure}

\begin{figure}[tb]
    \centering
    \includegraphics[width=\linewidth]{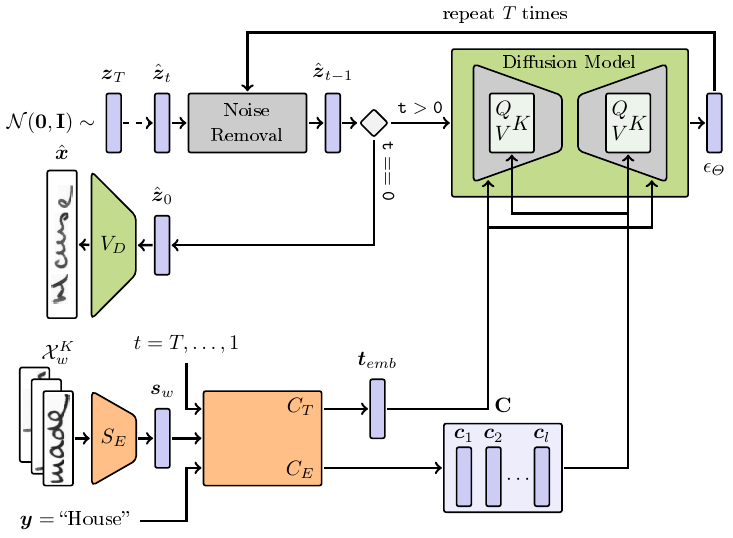}
    \caption{
        Overview of the proposed system when sampling an image given the desired string $\transcr{}$ and a set $\mathcal{X}_{w}^{K}$ of examples from the desired writer.
        Starting from random noise $\latentvec{z}{T} \sim \standardnormaldist$, the noise $\dm{\latentvec{z}{t}, \timestepemb, \contentemb}$ is predicted and removed.
        This procedure is repeated for $T$ timesteps to compute $\hat{\vec{z}}_0$ which is passed to the decoder part $\vaedecname$ of the VAE to obtain the final image $\hat{\vec{x}}$.}
    \label{fig:method-overview-sampling}
\end{figure}

Our proposed architecture is mainly based on the Wordstylist model~\cite{Nikolaidou2023_WSV}.
The generative model is a latent diffusion model which we describe in more detail in Sec.~\ref{subsec:method-ldm}.
In contrast to~\cite{Nikolaidou2023_WSV}, we add a dedicated model for the computation of the style embeddings, wich is discussed in Sec.~\ref{subsec:method-mae-style-encoder}.
Additionally, we extended the content encoder to not only encode the textual content but consider style information as well (cf.~Sec.~\ref{subsec:method-content-encoder}).
The overall architecture is depicted in figures~\ref{fig:method-overview-train} and~\ref{fig:method-overview-sampling}.

\subsubsection{Training}
\label{subsubsec:method-training}
For training, an image $\latentvec{x}{0}$ from writer $\wid{}$, its transcription $\transcr{}$ and a set $\mathcal{X}_{w}^{K}$ of examples from writer $\wid{}$ are given.
A style embedding vector $\writeremb{w}=\mae{\mathcal{X}_{w}^{K}}$ for the writer is computed by our style encoder and the latent representation of the image $\latentvec{z}{0}=\vaeenc{\latentvec{x}{0}}$ is computed by the VAE encoder.
Then, a random timestep $t \sim \mathcal{U}( 1, T)$ is drawn and the according amount of noise is added to the image by the forward diffusion process to obtain $\latentvec{z}{t}$.
$\transcr{}, \writeremb{w}$ and $\timestep$ are passed to the content encoder in order to generate a conditioning sequence $\contentembseq{l}=\contentenc{\transcr{}, \writeremb{w}}$ and timestep embedding $\timestepemb = \contentencts{\timestep, \writeremb{w}}$.
For better readability, we denote the $l$ embedding vectors $\contentvec_i \in \mathbb{R}^{\contentdim}$ of the sequence as the matrix $\contentemb \in \mathbb{R}^{\contentdim \times l}$.
Note that depending on the selected method for including the style conditioning (cf.\ Sec.~\ref{subsec:method-content-encoder}), the style conditioning may be contained in $\timestepemb$.
Given the latent $\latentvec{z}{t}$, timestep embedding $\timestepemb$ and the conditioning $\contentemb$, the diffusion model predicts the noise in the latent $\dm{\latentvec{z}{t}, \timestepemb, \contentemb}$ and the loss is computed as in Eq.~\ref{eq:dm-training-simple-loss}.
$\vaedecname$ is not needed during training.

\subsubsection{Sampling}
\label{subsubsec:method-sampling}
For generating a handwritten word image, a string for the textual conditioning $\transcr{}$ and a set $\mathcal{X}_{w}^{K}$ of writer examples for the style conditioning must be provided.
Like in training, a style embedding vector $\writeremb{w}=\mae{\mathcal{X}_{w}^{K}}$ is computed.
The reverse process starts at timestep $\timestep=T$ with the latent $\latentvec{z}{T} \sim \standardnormaldist$, conditioning $\contentemb=\contentenc{\transcr{}, \writeremb{w}}$ and timestep embedding $\timestepemb = \contentencts{\timestep, \writeremb{w}}$.
Noise prediction $\dm{\latentvec{z}{t}, \timestepemb, \contentemb}$ and removal of the noise (cf.\ Secs.~\ref{subsec:dm-sampling} and~\ref{subsec:dm-conditional-sampling}) from the latent is then repeated for $T$ timesteps during the reverse diffusion process in order to obtain $\hat{\vec{z}}_{0}$.
Finally, the generated image is computed by $\hat{\vec{x}} = \vaedec{\hat{\vec{z}}_{0}}$.

\subsection{Latent Diffusion Model}
\label{subsec:method-ldm}
For our LDM, we use the architecture from~\cite{Nikolaidou2023_WSV}.
The model is composed of a large-scale pre-trained VAE and a UNet for learning the reverse process~\cite{Rombach2022_HRI}.
While the conditional LDM proposed in~\cite{Rombach2022_HRI} was trained on large-scale datasets for generating natural images, the architecture needs to be modified to work with scarce data and lower resolution (height) of handwritten word images~\cite{Nikolaidou2023_WSV}.
We use the pre-trained VAE\footnote{We load the pre-trained weights from the Huggingface repository \url{https://huggingface.co/runwayml/stable-diffusion-v1-5}}~\cite{Rombach2022_HRI} keeping its weights frozen during training.
For the UNet, we followed~\cite{Nikolaidou2023_WSV} and reduced the number of Residual Blocks~\cite{He2016_DRL}, the number of attention heads and the internal feature dimensionality.
In order to maintain some spatial resolution\footnote{A height of $4$ pixels in the smallest feature map for input images with a height of $64$ pixels}, the latent representation of the VAE is downsampled only once within the UNet.

The VAE is unconditional and produces a latent space with only $1/8$ of each of the spatial dimensions of the image space.
The diffusion process is implemented in the latent space and parametrized by a conditional noise prediction $\dm{\latentvec{z}{t}, \timestepemb, \contentemb}$.
The architecture is based on an improved~\cite{Ho2020_DDP} UNet~\cite{Ronneberger2015_UNC} mainly composed of residual blocks~\cite{He2016_DRL} and self-attention blocks~\cite{Vaswani2017_AiA}.
Information about the timestep is included by adding the timestep embedding from the content encoder $\timestepemb = \contentencts{\timestep, \writeremb{w}}$ to the intermediate representation in the residual blocks.

To enable conditioning, we followed~\cite{Rombach2022_HRI} and replaced each of the self-attention layers by transformer blocks with one multi-headed self-attention layer and a multi-headed cross-attention layer (MHCA)~\cite{Vaswani2017_AiA} followed by an MLP\@.
Given the conditioning $\contentemb \in \mathbb{R}^{\contentdim \times l}$ from the content encoder (cf.\ Sec.~\ref{subsec:method-content-encoder}) and the intermediate representation of the UNet $\phi_i(\latentvec{z}{t}) \in \mathbb{R}^{N_i\times d_i}$ at layer $i$ with $N_i =w_i \cdot h_i$ being the flattened spatial dimensions and $d_i$ the feature dimension of the input feature map of that layer,
the MHCA with $H$ heads indexed with $h$ is computed between all spatial positions in the feature map and all positions in the conditioning sequence by
\begin{equation}
    \label{eq:method-mhca-qkv}
    \mat{Q}_h = \phi_i(\latentvec{z}{t}) \mat{W}^{(i)}_{Q_h} ,\hspace{0.5cm} \mat{K}_h = \contentemb^T \mat{W}^{(i)}_{K_h},\hspace{0.5cm} \mat{V}_h = \contentemb^T \mat{W}^{(i)}_{V_h}
\end{equation}
\begin{equation}
    \label{eq:method-mhca-head}
    \mat{H}_h = \text{softmax}\left( \frac{\mat{Q}_h\mat{K}_h^T}{\sqrt{d_{ca}}} \cdot \mat{V}_h \right)
\end{equation}
\begin{equation}
    \label{eq:method-mhca-out}
    \text{MHCA}\left(\phi_i(\latentvec{z}{t}), \contentemb \right) = \left[ \mat{H}_1,\dots,\mat{H}_H \right] \mat{W}^{(i)}_O
\end{equation}
with $\mat{W}^{(i)}_Q \in \mathbb{R}^{d_i \times d_{ca}}, \mat{W}^{(i)}_K \in \mathbb{R}^{\contentdim \times d_{ca}}, \mat{W}^{(i)}_V \in \mathbb{R}^{d_{ca} \times d_{ca}}, \mat{W}^{(i)}_O \in \mathbb{R}^{(d_{ca} \cdot H) \times d_i}$ and $d_{ca}$ being the internal dimension of the cross attention heads~\cite{Bishop2024_DLF}.

\subsection{Content Encoder}
\label{subsec:method-content-encoder}
\begin{figure}[tb]
    \centering
    \includegraphics[width=0.7\textwidth]{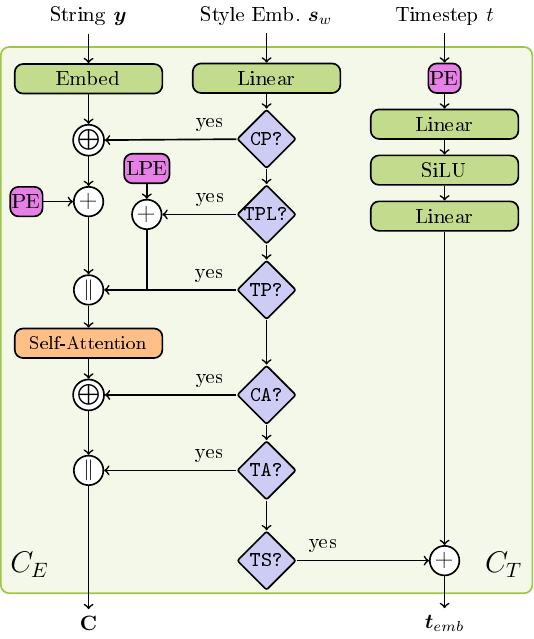}
    \caption{
        Computation of the content embedding $\contentemb=\contentenc{\transcr{}, \writeremb{w}}$ and the timestep embedding $\timestepemb = \contentencts{\timestep, \writeremb{w}}$ for different choices of incorporating the style embedding.
        \textit{PE} and \textit{LPE} denote positional encodings and learned positional embeddings, respectively.
        Operands are $+$ for elementwise addition, $\concat$ for concatenation of two vectors and $\append$ for appending a vector to the sequence.}
    \label{fig:method-content-encoder}
\end{figure}

As the diffusion model is conditioned on text as well as on style information, we extend the content encoder proposed in~\cite{Nikolaidou2023_WSV} to not only process the textual conditioning but also incorporate the style conditioning.
Both conditions, text and style, are closely related, i.e.\ every character has a specific appearance for some writer.
Therefore, we assume that incorporating the style information with the text conditioning before passing it to the UNet, might help improve generation performance.
The overall architecture of the content encoder for different ways of including the style embedding is depicted in Fig.~\ref{fig:method-content-encoder}.
The content encoder receives the string $\transcr{}=\transcrseq{k}$, $k\le l$ of the word to be generated, the style embedding $\writeremb{w}$ and the timestep $\timestep$ as input and computes the content embedding $\contentemb=\contentenc{\transcr{}, \writeremb{w}} \in \mathbb{R}^{\contentdim \times l}$ and the timestep embedding $\timestepemb = \contentencts{\timestep, \writeremb{w}} \in \mathbb{R}^{\timestepembdim}$.

\subsubsection{Text Embedding}
\label{subsubsec:method-text-embedding}
First, the given string is padded to fixed length $l$ and embedded into a sequence of character embeddings $\vec{y}_{emb_i} \in \mathbb{R}^{\characterembdim}$.
A sinusoidal positional embedding~\cite{Vaswani2017_AiA} is then added to the character embeddings:
\begin{equation}
    \label{eq:method-content-encoder-character-embedding}
    \left(\vec{y}_{emb_1}, \dots, \vec{y}_{emb_l}\right) = \left(\text{Embed}(y_1) + PE_{1}, \dots, \text{Embed}(y_l) + PE_{l}\right).
\end{equation}
The $\characterembdim$-dimensional positional encoding at position $pos$ and dimension $i$ is computed by
\begin{align}
    \label{eq:method-content-encoder-positional-encoding}
    PE_{(pos, 2i)} &= \sin(pos/10000^{2i/\characterembdim}) \\
    PE_{(pos, 2i+1)} &= \cos(pos/10000^{2i/\characterembdim}).
\end{align}
To allow the model to distribute information within the sequence, the final conditioning computed by a self-attention layer applied on the sequence of character embeddings:
\begin{equation}
    \label{eq:method-content-encoder-self-attention}
    \contentembseq{l} = \selfattention{\vec{y}_{emb_1}, \dots, \vec{y}_{emb_l}}.
\end{equation}

\subsubsection{Style Embedding}
\label{subsubsec:method-style-embedding}
In~\cite{Nikolaidou2023_WSV}, the style embedding vector is passed with the same dimensionality as the time step embedding for the diffusion model and then added to it.
The obtained embedding is directly passed to the residual blocks of the UNet.
However, by this way of including the style conditioning, there is only interaction between features influenced by the style embedding and the character embeddings in the cross-attention of the transformer blocks of the UNet.

In order to allow the model to relate style and text in advance, we propose different approaches for including the style embedding.
As a first step, we always use a learnable linear projection $\mat{W}_s \in \mathbb{R}^{\writerembdim \times d_x}$ to allow small adjustments of the provided style embedding as well as to adjust the dimensionality of the vector for later steps:
\begin{equation}
    \label{eq:mathod-content-encoder-style-linear-proj}
    \hat{\vec{s}}_{w} = \mat{W}_s\writeremb{w}, \hspace{1cm} \writeremb{w} \in \mathbb{R}^{\writerembdim},
\end{equation}
where $d_x$ is set to either the dimensionality of the timestep embedding $\timestepembdim$ or the character embedding $\characterembdim$.

To allow the model to not only exchange character information with the self-attention in the content encoder but also style information, we propose to append the style embedding to the sequence of character embedding vectors before the self-attention layer:
\begin{equation}
    \label{eq:method-content-encoder-tp}
    \contentembseq{l+1}_{\text{\tokenpreattention}} = \selfattention{\vec{y}_{emb_1}, \dots, \vec{y}_{emb_l}, \hat{\vec{s}}_{w}}.
\end{equation}
Optionally a learnable positional embedding can be added to this token:
\begin{equation}
    \label{eq:method-content-encoder-tpl}
    \contentembseq{l+1}_{\text{\tokenpreattentionlemb}} = \selfattention{\vec{y}_{emb_1}, \dots, \vec{y}_{emb_l}, \hat{\vec{s}}_{w} + LPE}.
\end{equation}
We refer to these options as \tokenpreattention{} and \tokenpreattentionlemb, respectively.
Another option is the concatenation (denoted by $\oplus$) of the style embedding vector to all character embedding vectors before the self-attention (\concatpreattentnion):
\begin{equation}
    \label{eq:method-content-encoder-cp}
    \contentembseq{l}_{\text{\concatpreattentnion}} = \selfattention{\vec{y}_{emb_1}\oplus\hat{\vec{s}}_{w}, \dots, \vec{y}_{emb_l}\oplus\hat{\vec{s}}_{w}}.
\end{equation}
Furthermore, the style embedding can be appended to the sequence after the self-attention (\tokenafterattention):
\begin{align}
    \label{eq:method-content-encoder-ta}
    (\hat{\contentvec}_1,\dots,\hat{\contentvec}_{l}) &= \selfattention{\vec{y}_{emb_1}, \dots, \vec{y}_{emb_l}} \\
    (\contentvec_1,\dots,\contentvec_{l+1})_{\text{\tokenafterattention}} &= (\hat{\contentvec}_1,\dots,\hat{\contentvec}_{l},\hat{\vec{s}}_{w}).
\end{align}
While not allowing style information to be included in the whole embedding sequence, this inclusion enables the transformer blocks of the Unet to attend specifically on the style embedding vector.
We also explore the concatenation of the style embedding to every vector of the sequence after the self-attention (\concatafterattention):
\begin{equation}
    \label{eq:method-content-encoder-ca}
    \contentembseq{l}_{\text{\concatafterattention}} = (\hat{\contentvec}_1\oplus\hat{\vec{s}}_{w},\dots,\hat{\contentvec}_{l}\oplus\hat{\vec{s}}_{w}).
\end{equation}
This inclusion provides style information in every vector of the input sequence for the cross-attention while not influencing the self-attention on the character embedding sequence.
We also experiment with the approach of adding the style embedding to the timestep embedding (\addtotimestep).
However, we use a learnable linear projection of the given style embedding, as for our method the given vector might have a different dimensionality than the time step embedding:
\begin{equation}
    \label{eq:method-content-encoder-ts}
    \timestepemb = \text{Linear}(\text{SiLU}(\text{Linear}(PE(t)))) + \hat{\vec{s}}_{w}.
\end{equation}
For all inclusion methods, the computed content embedding vectors $\contentvec_i$ have the same dimensionality $\contentdim = \characterembdim$ as the character embeddings except for the concatenation-based inclusions \concatpreattentnion{} and \concatafterattention{} where $\contentdim = 2\characterembdim$.

\subsection{MAE Style Encoder}
\label{subsec:method-mae-style-encoder}
In~\cite{Nikolaidou2023_WSV}, the style embedding is a learned embedding vector per writer.
This however, inhibits the generation in styles from writers not seen during training.
We overcome this limitation by using an additional model for computing style embedding vectors, given some example images from a writer like~\cite{Kang2020_GCC}.
For computation of the style embeddings, we use a modified version of the masked autoencoder (MAE) proposed by Souibgui et al.~\cite{Souibgui2023_TDs}.
We assume that the unmasking task pushes the model to encode information about the strokes in the embeddings.
Additionally, a major advantage is that the model is trained in a self-supervised fashion requiring no annotations of the writer IDs.
In this way, we avoid being restricted to labeled training data.

The MAE is composed of an encoder and a decoder, denoted by $\styleencname$ and $\maedecname$.
The encoder $\styleencname$ is a vanilla vision transformer (ViT)~\cite{Dosovitskiy2021_IiW}.
A given word image $\img{}$ is divided into a set of $N$ non-overlapping patches $\patchify{X}{\img{}}{N}$ which are then encoded by a learned linear projection.
Note that for the explanation of the MAE the subscript does not refer to the diffusion timestep but is used for indexing the patches and their encodings of the original image.
Positional information is considered by the concatenation with a positional embedding per patch.
The latent representation $\patchify{Z}{\vec{z}}{N}=\styleenc{\img{}_p}$ is obtained by processing these embeddings by a sequence of transformer blocks~\cite{Souibgui2023_TDs}.
Each transformer block is composed of a Layer Norm~\cite{Ba2016_LN} and a Multi-Head Self-Attention, followed by another Layer Norm and a Multi-Layered Perceptron.
A decoder $\maedecname$ with the same structure as the encoder followed by a linear projection is used to predict flattened patches $\tilde{\img{}}_p = \maedec{\vec{z}_p}$ out of the latent representation.

During training of a MAE~\cite{He2022_MAA}, $r$ percent of the patches are removed randomly.
Given a masking ratio $r\in (0, 1)$ a random binary mask $\mathcal{M}=\left\{0, 1 \right\}^N$ containing $N\cdot r$ ones is computed per image.
We denote the set of masked patches as $\mathcal{X}_m = \left\{ \img{}_{p_i} | \mathcal{M}_i = 1 \right\}$ and the set of visible patches as $\mathcal{X}_v = \left\{ \img{}_{p_i} | \mathcal{M}_i = 0 \right\}$.
For the visible patches, the latent representations $\vec{z}_{p_i}=\styleenc{\mathcal{X}_v}$ are computed.
Before decoding, a learnable mask token is inserted at the positions of the masked patches.
The encoder and decoder are then trained to predict the masked patches from the context provided by the embeddings from the non-masked patches:
\begin{equation}
    \label{eq:method-l2-loss}
    \mathcal{L}_{\text{mask}} = \sum_{i=1}^{N} \mathcal{M}_i * \norm{\maedec{\styleenc{\mathcal{X}_v}}_i - \img{}_{p_i}}^2.
\end{equation}

\cite{Souibgui2023_TDs} train their MAE in a multitask fashion for prediction of masked patches, noise removal and deblurring.
For each task, they use a dedicated linear projection in the encoder and in the decoder.
In their ablation study, Souibgui et al.\ show that focusing only on masking for pre-training works best when using the writer embedding for handwritten text recognition.
While this evaluation shows, that textual information must be encoded by the model, we need the embeddings to encode stylistic information.
For a good reconstruction of the input image, however, the model must have encoded stylistic information in addition to the textual information.
Therefore, we decided to train our MAE with the unmasking objective only.

After training the MAE, writer embeddings can be computed for any given set $\mathcal{X}_{\wid{}}^{K}$ of example images from that writer where $K$ is the number of images.
Using the encoder of the MAE, the latent representations for all $N$ patches are computed and then averaged to obtain a global embedding for each image.
The style information of all $K$ examples from the writer is then combined by averaging the embedding vectors of the images resulting in a single embedding vector for the writer:
\begin{equation}
    \label{eq:method-style-encoding}
    \styleenc{\wid{}} = \frac{1}{K * N} \sum_{i=1}^{K}\sum_{j=1}^{N} \styleenc{\img{i}_{p_j}}.
\end{equation}

\subsection{Semi-Supervised Training}
\label{subsec:method-semi-supervised-training}
While in theory, the approach described so far should be able to generate handwritten word images for unseen writers, differences between the styles seen during training and the style requested for generation might deteriorate the generation quality.
However, this is exactly the scenario in which our model should help reducing the lack of annotated data by conditional generation of new samples.
Since our approach does not need explicitly labeled writer annotations but only a few samples known to be written by the desired writer (e.g.\ words from the same row or page), we assume that we can compute style embeddings for the new writers.
Thereby, we can utilize the data for training our model in a semi-supervised fashion.
By providing fully labeled samples (writer and transcription), our model is trained to incorporate the given conditioning on text and calligraphic style.
Hereby the model can learn the correspondences between the text and style conditioning (e.g.\ the appearance of the characters for a given style conditioning).
Incorporating the partially labeled (style conditioning only\footnote{Note that in theory the model could be trained in a semi-supervised fashion without both, transcriptions and style embeddings, for the new dataset. We leave this open for future work.}) samples allows the model to explore areas of the latent space which might extend the latent space made up by the fully labeled dataset.
Therefore, we assume that utilizing these examples should improve generation quality for the new writers.
Technically, we replace the sequence of embedding vectors for the text conditioning with the same mask tokens as for classifier-free guidance.

    \section{Experiments}
\label{sec:experiments}
In this section we evaluate our proposed HTG system.
As we believe that generation of training images for a downstream model is a promising application of HTG, we focus on this scenario for our evaluation.
Particularly, the transfer from one dataset to unseen writers and words is of high interest for this application.
Therefore, we assume to have one fully labeled data set and a set of handwritten word images where can only assume a writer per set of images but do not know any transcription.
We evaluate generation performance for both datasets.

\subsection{Datasets}
\label{subsec:experiments-datasets}
For our experiments, we use the IAM-database~\cite{Marti2002_IdE} and the RIMES-database~\cite{Grosicki2009_I2H} on word level.
Both datasets contain examples of modern handwriting of English (IAM) and French (RIMES) words from multiple writers.
We only consider words with 2--7 latin characters ([a-zA-Z]) like~\cite{Nikolaidou2023_WSV}\footnote{While not explicitly mentioned in their paper, Kang et al.~\cite{Kang2020_GCC} have the same limitations in the annotation files published with their implementation.}.
Following recent works~\cite{Kang2020_GCC,Nikolaidou2023_WSV,Mattick2021_SIH}, we use the RWTH Aachen split for IAM\@ and use the combined training and validation sets for training the models.
More detailed statistics about dataset size, number of writers, lexicon size and OOV samples are provided in Tab.~\ref{tab:experiments-datasets}.
Despite the fact that the datasets were created in different languages, there is around $20\,\%$ overlap of the lexicon of IAM-train and the one from RIMES\@.
Note that for both datasets, the writers of the test set do not overlap with the writers from their respective train sets.
\begin{table}[tb]
    \caption{Number of samples, number of writers and lexicon size of IAM~\cite{Marti2002_IdE} and RIMES~\cite{Grosicki2009_I2H} after filtering.
    Additionally, the portion of out-of-vocabulary samples in the test set compared to the train set (OOV) and compared to IAM-train (OOV IAM-train) are presented.}
    \label{tab:experiments-datasets}
    \centering
    \begin{tabular}{l|l|r|r|r|r|r}
        \toprule
        Dataset & Partition & \#Samples & \#Writers & \#Lexicon & OOV $\left[ \% \right]$ & OOV IAM-train $\left[ \% \right]$ \\
        \midrule
        IAM     & train     & $44405$   & $338$     & $4916$    & --                      & --                                \\
        IAM     & test      & $18436$   & $161$     & $3332$    & $9.42$                  & --                                \\
        RIMES   & train     & $34468$   & $1343$    & $1740$    & --                      & $81.19$                           \\
        RIMES   & test      & $5182$    & $179$     & $658$     & $2.43$                  & $80.57$                           \\
        \bottomrule
    \end{tabular}
\end{table}

To simplify training in batches, we resize all images to a height of $64$ pixels preserving the aspect ratio, similar to~\cite{Nikolaidou2023_WSV,Nikolaidou2024_DTC}.
Our desired width is $256$ pixels.
If the resulting image is smaller than this, we pad the image to the right with white pixels.
Larger images are scaled to fit $256 \times 64$ pixel.
While the scaling changes the aspect ratio, there are only little distortions due to the limitation to 7 characters~\cite{Nikolaidou2023_WSV}.

For our experiments, we assume to have one fully labeled dataset (i.e.\ transcriptions and writer IDs) and one \textit{unseen} dataset, where we can only infer which samples belong to the same writer (e.g.\ assume all words from one page are written by the same person).
We use IAM as our fully labeled dataset for the training of every diffusion model.
For experiments with semi-supervised training we add examples from RIMES without transcriptions.

\subsection{Evaluation Measures}
\label{subsec:experiments-evaluation}
The assessment of generated images is a challenging problem, especially for the generation of handwriting.
Despite the overall visual quality, the generated image has to show the correct text in the desired calligraphic style.
Besides a qualitative assessment by humans, the Fréchet Inception Distance (FID)~\cite{Heusel2017_GTT} is commonly used for evaluation of generated natural images.
The FID measures the difference in the distributions of feature vectors extracted by an InceptionV3 model~\cite{Szegedy2016_RIA}.
Since this model was trained on natural images~\cite{Deng2009_Ils}, it is not well suited for handwritten word images~\cite{Kang2020_GCC,Nikolaidou2023_WSV}.
Furthermore, the correctness of the content is not considered.
Instead, we employ different evaluation measures with a focus on the downstream task of handwriting generation.
A common approach is the usage the generated data for training a handwritten text recognition (HTR) model~\cite{Elanwar2024_Gan}.
The HTR model is then tested on the original test set and the character error rate (CER) is computed, in the following referred to as \textit{CER-train}.
For our experiments we replicate the train set (i.e.\ the same text and style conditioning pairs like in the original train set).
CER-train therefore does not capture diversity well, but all the more fidelity and correctness with respect to the conditioning.

Additionally, we train an HRT model on the original training images of IAM\@.
We use our diffusion model to replicate the test set of IAM and compute the difference between the CER of the HTR model on the generated samples and the CER on the original test samples.
Since the writers from IAM train and test are disjoint, all styles can be considered unseen.
Furthermore, we distinguish between in-vocabulary (IV) and out-of-vocabulary (OOV) examples.
In the IV case, the writers are unseen but images of the strings were already generated during training.
We refer to this metric as \textit{Diff-IV}.
High values imply that the overall generation quality and correctness of the content is poor and the word is not readable.
Large negative values indicate that the textual content is correct but the model overfits the styles of the train samples, making it easier for the HTR model to predict the correct characters.
Therefore, values close to $0$ are appreciable, indicating that the textual content is correct and the writing style is similarly difficult to recognize compared to the original test set.

Since in the OOV case writer and style are unseen, it is unclear whether the generation of the style, the generation of the text or the generalization capabilities of the HTR model are the cause of bad results.
We therefore only consider Diff-IV as a second metric for experiments on the IAM dataset besides CER-train.

\subsection{Implementation Details}
\label{subsec:experiments-implementation-details}

\subsubsection{Style Encoder}
\label{subsubsec:experiments-setup-style-encoder}
We first train our style encoder either on IAM or on both, IAM and RIMES, using the same setup as~\cite{Souibgui2023_TDs} for their HTR pre-training.
The model is trained on RGB images with a resolution of $64 \times 256$ pixel.
These are split into patches of $8 \times 8$ pixel of which $75\,\%$ are randomly masked.
The encoder is comprised of $6$ layers with $8$ attention heads (same for decoder) and computes patch embeddings with $768$ dimensions.
The model is trained for $100$ epochs with a batch size of $64$ using AdamW~\cite{Loshchilov2019_DWD} with a learning rate of $1.5\cdot 10^{-4}$ with cosine decay, $3$ epochs warmup, momentums $\beta_1=0.9$ and $\beta_2=0.95$, and a weight decay of $0.05$.
After training the style encoder, we compute embeddings for all writers.
The embeddings are averaged over $10$ random example images each.
This choice is motivated by the assumption that it is easy to collect $10$ examples for new writers if one or two handwritten sentences are available.
For training the diffusion model, we repeat this sampling $100$ times per writer as an augmentation in order to better explore the space of writer embeddings.

\subsubsection{Latent Diffusion Model}
\label{subsubsec:experiments-setup-ldm}
We train our diffusion model similar to~\cite{Nikolaidou2023_WSV}.
However, instead of passing a writer ID to the model, we randomly draw one of the $100$ embeddings per writer every iteration.
We use a linear variance schedule from $\beta_1=10^{-4}$ to $\beta_T=0.02$ with $T=1000$ steps for the forward diffusion process.
The model is trained for $1000$ epochs with a batch size of $224$.
As optimizer, we use AdamW~\cite{Loshchilov2019_DWD} with a learning rate of $10^{-4}$, momentums $\beta_1=0.9$ and $\beta_2=0.999$, and a weight decay of $0.01$.
Weights are updated based on an exponential moving average $w_{i-1} * \gamma + (1 - \gamma) * w_{i}$ with $\gamma=0.995$.
For the experiments with classifier-free guidance, we independently drop writer and text conditioning with a probability of $p=0.1$.
Like~\cite{Nikolaidou2023_WSV}, we do not apply any augmentations to the training images in order to not distort the writing styles.
Images are sampled using UniPC~\cite{Zhao2023_UUP} with 50 noise steps and afterward cropped to the area containing text.

\subsubsection{Recognition Model}
\label{subsubsec:experiments-setup-htr}
We use the HTR model presented by Retsinas et al.~\cite{Retsinas2022_BPH} for our evaluation.
The model comprises a convolutional backbone followed by two heads, one recurrent head consisting of three bidirectional LSTM layers and one shortcut head only consisting of a 1D convolutional layer.
The latter head serves as an auxiliary component during training and is discarded when testing.
The model is trained using Connectionist Temporal Classification~\cite{Graves2006_Ctc} on images padded to a fixed size of $256 \times 64$ pixel.
We use AdamW~\cite{Loshchilov2019_DWD} for optimization with a learning rate of $10^{-3}$, momentums $\beta_1=0.9$ and $\beta_2=0.999$, and a weight decay of $5\cdot 10^{-5}$.
We train the model for $240$ epochs and divide the learning rate by $10$ after $120$ and $180$ epochs.
Tab.~\ref{tab:experiments-htr-ref} reports the reference performance of the HTR model on the test sets of IAM and RIMES when trained on the respective original train sets.
\begin{table}
    \caption{Reference CER $\left[ \% \right]$ of the HTR model on our IAM and RIMES test partitions (all).
    Additionally, we report results on in-vocabulary (IV) and out-of-vocabulary (OOV) test samples.}
    \label{tab:experiments-htr-ref}
    \centering
    \begin{tabular}{L{6em}|R{3em}|R{3em}|R{3em}}
        \toprule
        Dataset & all  & IV   & OOV   \\
        \midrule
        IAM     & 6.01 & 5.18 & 11.36 \\
        RIMES   & 2.47 & 1.58 & 25.04 \\
        \bottomrule
    \end{tabular}
\end{table}

\subsection{Style Inclusion and Classifier-free Guidance}
\label{subsec:experiments-si-cfg-gs}
First, we compare the different approaches for including the style conditioning using only IAM for training the style encoder and the diffusion model.
For every style inclusion, we trained one model without dropping of labels during training ($p_{\text{uncond}}=0$) and sampled images without guidance in these cases.
Additionally, we trained models with $p_{\text{uncond}}=0.1$ and sampled images without guidance as well as with guidance scales $w_{gs} \in \left\{ 2, 3, 4, 5, 6, 7 \right\}$.
Both models were trained using the same style embeddings.
Results are shown in Tab.~\ref{tab:experiments-iam-only-cer-train} and~\ref{tab:experiments-iam-only-cer-iv-diff}.
\begin{table}[tb]
    \caption{Results of DMs trained with different style inclusions and CFG configurations on IAM. Sampling was done using different guidance scales.}
    \begin{subfigure}{\textwidth}
        \caption{CER-train $\left[ \% \right]$}
        \label{tab:experiments-iam-only-cer-train}
        \centering
        \begin{tabular}{L{5em}|C{4em}|R{3em}|R{3em}|R{3em}|R{3em}|R{3em}|R{3em}}
            \toprule
            Guidance   & $p_\text{uncond}$ & \tokenpreattention{} & \tokenpreattentionlemb{} & \concatpreattentnion{}   & \tokenafterattention{}   & \concatafterattention{}   & \addtotimestep{} \\
            \midrule
            No CFG     & $0.0$             & 8.68                 & 8.57                     & 8.74                   & 8.48                   & 9.06                    & \textbf{8.08}    \\
            \midrule
            Unguided   & $0.1$             & 9.10                 & 9.18                     & 9.33                   & 9.19                   & 9.49                    & \textbf{8.85}    \\
            $w_{gs}=2$ & $0.1$             & 8.17                 & 8.11                     & 8.51                   & 8.70                   & 8.24                    & \textbf{7.84}    \\
            $w_{gs}=3$ & $0.1$             & 7.62                 & 8.05                     & 7.98                   & 8.02                   & 8.08                    & \textbf{7.54}    \\
            $w_{gs}=4$ & $0.1$             & 7.78                 & 7.77                     & 8.18                   & 7.72                   & 7.93                    & \textbf{7.57}    \\
            $w_{gs}=5$ & $0.1$             & 7.56                 & 7.98                     & 7.60                   & 7.84                   & 7.82                    & \textbf{7.39}    \\
            $w_{gs}=6$ & $0.1$             & \textbf{7.54}        & 7.64                     & 7.73                   & 7.66                   & 7.72                    & 7.69             \\
            $w_{gs}=7$ & $0.1$             & 7.59                 & 7.61                     & 8.07                   & 7.79                   & 7.70                    & \textbf{7.56}    \\
            \bottomrule
        \end{tabular}
    \end{subfigure}
    \begin{subfigure}{\textwidth}
        \caption{Diff-IV (in percentage points)}
        \label{tab:experiments-iam-only-cer-iv-diff}
        \centering
        \begin{tabular}{L{5em}|C{4em}|R{3em}|R{3em}|R{3em}|R{3em}|R{3em}|R{3em}}
            \toprule
            Guidance   & $p_\text{uncond}$ & \tokenpreattention & \tokenpreattentionlemb & \concatpreattentnion   & \tokenafterattention   & \concatafterattention   & \addtotimestep \\
            \midrule
            No CFG     & $0$               & 3.83               & 3.50                   & 3.72                 & 3.61                 & 5.94                  & \textbf{2.66}  \\
            \midrule
            Unguided   & $0.1$             & 5.30               & 5.11                   & 5.92                 & 5.52                 & 6.14                  & \textbf{3.83}  \\
            $w_{gs}=2$ & $0.1$             & -0.65              & -0.83                  & -0.66                & -0.60                & \textbf{-0.24}        & -1.11          \\
            $w_{gs}=3$ & $0.1$             & -1.70              & -1.72                  & -1.55                & -1.51                & \textbf{-1.17}        & -1.83          \\
            $w_{gs}=4$ & $0.1$             & -2.06              & -2.14                  & -1.96                & -1.92                & \textbf{-1.64}        & -2.10          \\
            $w_{gs}=5$ & $0.1$             & -2.32              & -2.27                  & -2.13                & -2.05                & \textbf{-1.73}        & -2.28          \\
            $w_{gs}=6$ & $0.1$             & -2.39              & -2.39                  & -2.17                & -2.19                & \textbf{-1.78}        & -2.27          \\
            $w_{gs}=7$ & $0.1$             & -2.42              & -2.40                  & -2.20                & -2.24                & \textbf{-1.88}        & -2.27          \\
            \bottomrule
        \end{tabular}
    \end{subfigure}
\end{table}
Adding the style embedding vector to the timestep embedding (\addtotimestep{}) achieves best results in CER-train except for $w_{gs}=6$ where \tokenpreattention{} performs slightly better.
While dropping of the conditioning during training harms generation performance when sampling is unguided, classifier-free guidance helps when sampling with a guidance-scale $w_{gs} \ge 2$.
The largest increase in performance can be observed when increasing the guidance scale up to $w_{gs}=3$.
Increasing the guidance scale further up to $w_{gs}=7$ mostly results in small improvements only.
For some cases, also a slight decrease in performance can be observed, as an increased guidance scale during sampling is known to compromises sample quality for diversity~\cite{Ho2022_CFD}.

For all style inclusions, training the DM without CFG and unguided sampling result in high values of Diff-IV, indicating poor generation performance.
Training the DM with CFG and sampling with guidance-scales $w_{gs} \ge 2$ results in negative values of Diff-IV\@.
Therefore, we can assume that in these cases the generation overfits to the styles of the train set.
For guided sampling, the concatenation of the style embedding to every vector of the sequence after the self-attention (\concatafterattention{}) achieves the best results.

\subsection{Semi-supervised Training}
\label{subsec:experiments-generalization}
Up to now, only the generation of training and test examples from the same dataset were considered where a similar distribution can be assumed.
However, the more interesting application of DMs for HTG lies in the generation of examples of other, previously unseen (potentially unlabeled) datasets.
For this scenario, we consider the RIMES dataset to have no transcriptions available.
We use the writer labels but argue, that these can be inferred easily as long as information about the pages a word was taken from is provided~\footnote{
    In this case one writer per page would be a reasonable assumption.
    With an average of $26$ images per writer, RIMES has a realistic amount of samples per writer for this procedure.}.

For reference, we trained a DM on RIMES with transcriptions and writer labels.
The writer embeddings used in this experiment were computed by an MAE trained on examples from RIMES\@.
Additionally, we conducted an experiment where we combined IAM and RIMES for training the DM on both datasets with labels.
In order to assess the influence of the training data for the MAE, we used MAEs trained on the individual datasets and the combined one for the computation of writer embeddings.
Since training DMs is computationally expensive, we only trained DMs with the style inclusion \addtotimestep{} which gave the best results in the experiments using only the IAM database.
The reference results of CER-train for RIMES are presented in Tab.~\ref{tab:experiments-RIMES-only-cer-train}.
\begin{table}[tb]
    \caption{CER-train $\left[ \% \right]$ for RIMES of DMs trained with different fully labeled datasets (row \enquote{DM DS}) and CFG with $p_\text{uncond}=0.1$.
    MAEs were trained on different datasets (row \enquote{MAE DS}), too.
    Sampling was done using different guidance scales.}
    \label{tab:experiments-RIMES-only-cer-train}
    \centering
    \begin{tabular}{L{5em}|R{4em}|R{4em}|R{4em}|R{4em}}
        \toprule
        DM DS & RIMES & \multicolumn{3}{c}{Both} \\
        \midrule
        MAE DS     & RIMES & RIMES         & IAM           & Both \\
        \midrule
        Unguided   & 4.16  & \textbf{3.93} & 3.95          & 4.12 \\
        $w_{gs}=2$ & 3.78  & \textbf{3.52} & 3.72          & 3.99 \\
        $w_{gs}=3$ & 3.78  & 3.58          & \textbf{3.46} & 3.58 \\
        $w_{gs}=4$ & 3.60  & \textbf{3.43} & 3.74          & 3.84 \\
        $w_{gs}=5$ & 3.42  & \textbf{3.25} & 3.72          & 3.77 \\
        \bottomrule
    \end{tabular}
\end{table}
Similar to the experiments in Sec.~\ref{subsec:experiments-si-cfg-gs}, we achieve performance improvements by guided sampling.
However, not in all cases an increase of the guidance scale is helpful.
While the models parameters were already reduced compared to~\cite{Rombach2022_HRI} (cf.~Sec.~\ref{subsec:method-ldm}), increasing the amount of training data mostly results in a slightly better performance indicating that the model capacity is still not fully utilized.
Interestingly, the choice of the training dataset for the MAE has only very little influence on the results.

As a first experiment for the generation of RIMES as an unseen dataset without transcriptions, we evaluate the generalization capability of the DMs trained only on IAM\@.
For the writer embeddings for RIMES we use the same MAE as before which was only trained on IAM samples.
The results are shown in Tab.~\ref{tab:experiments-iam-only-cer-train-RIMES}.
\begin{table}[tb]
    \caption{CER-train for RIMES $\left[\% \right]$ as a new dataset.
    For reference: We achieved a CER-train up to 3.25\,\% for RIMES when including fully labeled RIMES during training.}
    \begin{subfigure}{\textwidth}
        \caption{CER-train for RIMES $\left[\% \right]$ of DMs \textbf{trained only on IAM} with different style inclusions and CFG configurations.
        Sampling was done using different guidance scales.}
        \label{tab:experiments-iam-only-cer-train-RIMES}
        \centering
        \begin{tabular}{L{5em}|C{4em}|R{3em}|R{3em}|R{3em}|R{3em}|R{3em}|R{3em}}
            \toprule
            Guidance   & $p_\text{uncond}$ & \tokenpreattention & \tokenpreattentionlemb & \concatpreattentnion   & \tokenafterattention   & \concatafterattention   & \addtotimestep \\
            \midrule
            No CFG     & $0$               & 10.48              & \textbf{10.30}         & 10.47                & 11.25                & 12.49                 & 11.83          \\
            \midrule
            Unguided   & $0.1$             & 8.92               & \textbf{8.66}          & 10.07                & 8.73                 & 9.11                  & 9.30           \\
            $w_{gs}=2$ & $0.1$             & 9.24               & \textbf{9.01}          & 11.17                & 9.13                 & 10.42                 & 9.58           \\
            $w_{gs}=3$ & $0.1$             & 9.92               & 9.17                   & 10.59                & \textbf{8.99}        & 11.31                 & 10.19          \\
            $w_{gs}=4$ & $0.1$             & 9.96               & \textbf{9.23}          & 11.19                & 9.48                 & 11.07                 & 9.58           \\
            $w_{gs}=5$ & $0.1$             & 9.40               & \textbf{9.32}          & 11.73                & 9.56                 & 11.55                 & 9.86           \\
            \bottomrule
        \end{tabular}
    \end{subfigure}
    \begin{subfigure}{\textwidth}
        \caption{CER-train $\left[\% \right]$ for RIMES of DMs \textbf{trained on fully labeled IAM and RIMES without transcriptions} using different style inclusions and CFG with $p_\text{uncond}=0.1$.
        MAEs were trained on different datasets (column \enquote{MAE DS}).}
        \label{tab:experiments-semi-cer-train-RIMES}
        \centering
        \begin{tabular}{C{4em}|L{5em}|R{3em}|R{3em}|R{3em}|R{3em}|R{3em}|R{3em}}
            \toprule
            MAE DS                & Guidance   & \tokenpreattention & \tokenpreattentionlemb & \concatpreattentnion & \tokenafterattention   & \concatafterattention   & \addtotimestep \\
            \midrule
            \multirow{5}{*}{IAM}  & Unguided   & 7.52               & 8.65                   & 7.03                 & \textbf{6.88}        & 7.94                  & 7.15           \\
            & $w_{gs}=2$ & 6.96               & 7.39                   & 7.32                 & 7.08                 & 8.43                  & \textbf{6.61}  \\
            & $w_{gs}=3$ & 7.78               & \textbf{6.36}          & 8.23                 & 7.41                 & 8.48                  & 7.29           \\
            & $w_{gs}=4$ & 7.82               & \textbf{6.28}          & 7.66                 & 7.72                 & 8.67                  & 7.08           \\
            & $w_{gs}=5$ & 8.21               & \textbf{6.60}          & 7.63                 & 7.69                 & 8.69                  & 6.91           \\
            \midrule
            \multirow{5}{*}{Both} & Unguided   & 9.74               & 9.19                   & \textbf{6.95}        & 8.53                 & 7.24                  & 7.17           \\
            & $w_{gs}=2$ & 7.57               & 6.90                   & 7.16                 & 7.88                 & 6.68                  & \textbf{6.65}  \\
            & $w_{gs}=3$ & 7.05               & 7.42                   & 7.23                 & 7.99                 & 7.17                  & \textbf{7.04}  \\
            & $w_{gs}=4$ & 6.83               & 6.69                   & 7.40                 & 8.50                 & 7.17                  & \textbf{6.63}  \\
            & $w_{gs}=5$ & 6.80               & \textbf{6.48}          & 7.12                 & 8.60                 & 7.04                  & 6.96           \\
            \bottomrule
        \end{tabular}
    \end{subfigure}
\end{table}
Compared to the reference in Tab.~\ref{tab:experiments-RIMES-only-cer-train} when the DM is trained on labeled RIMES or a combination of IAM and RIMES, we can observe a major drop in performance for all style inclusions and guidance scales.
This shows that the model is by default not well suited for the generation of examples with a slightly different appearance but mostly different distribution of words (cf.~Tab.~\ref{tab:experiments-datasets}).
In contrast to the generation of examples from IAM, increasing the guidance scale $w_{gs}$ leads to deterioration of the results.

In order to mitigate the performance drop, we use a semi-supervised approach for training the DM (cf.~Sec.~\ref{subsec:method-semi-supervised-training}).
Additionally, we investigate whether it is beneficial if the MAE for computing the writer embeddings has seen examples from the new dataset during training.
This is possible since the MAE does not require any labels for training.
As can be seen in Tab.~\ref{tab:experiments-semi-cer-train-RIMES}, semi-supervised training of the DM leads to an increase in performance for all style inclusions and guidance scales, except for two cases using unguided sampling.
In most cases, using guided sampling is beneficial.
However, no continuous improvement in the results can be observed with an increase in the guidance scale as in the experiments with IAM\@.
On average, the best results were obtained with $w_{gs}=2$.
We assume that this is caused by the model being more uncertain about the conditioning as no textual labels were given for RIMES during training.
When the MAE was trained only on IAM, the best results were obtained using \tokenpreattentionlemb{}, except for unguided sampling and sampling with $w_{gs}=2$ where \tokenafterattention{} and \addtotimestep{} performed better.
In the experiments with the MAE trained on both datasets, the DM using \addtotimestep{} performs best in most cases, however achieving no better results than the best DM where the MAE was trained only on IAM\@.
For \concatpreattentnion{} and \concatafterattention{}, a performance improvement for all guidance scales can be observed when the MAE was trained on both datasets.
Only for \tokenafterattention{}, training the MAE on both datasets resulted in inferior performance for all guidance scales.

As we expect to introduce some noise by the semi-supervised training since around 50\,\% of the training data have no text conditioning, we evaluate CER-train for IAM as well.
In the results shown in Tab.~\ref{tab:experiments-semi-cer-train-iam}, we can observe the expected decrease in performance which however is very small.
\begin{table}[tb]
    \caption{CER-train for IAM $\left[\% \right]$ of DMs trained on fully labeled IAM and RIMES without transcriptions using different style inclusions and CFG with $p_\text{uncond}=0.1$.}
    \label{tab:experiments-semi-cer-train-iam}
    \centering
    \begin{tabular}{C{4em}|L{5em}|R{3em}|R{3em}|R{3em}|R{3em}|R{3em}|R{3em}}
        \toprule
        MAE DS                & Guidance   & \tokenpreattention & \tokenpreattentionlemb & \concatpreattentnion & \tokenafterattention   & \concatafterattention   & \addtotimestep \\
        \midrule
        \multirow{5}{*}{IAM}  & Unguided   & 9.73               & 9.42                   & 9.68                 & 9.61                 & 10.31                 & \textbf{8.90}  \\
        & $w_{gs}=2$ & 8.73               & 8.37                   & 8.60                 & 8.47                 & 8.59                  & \textbf{8.15}  \\
        & $w_{gs}=3$ & 8.26               & 8.17                   & 8.16                 & 8.21                 & 8.45                  & \textbf{7.78}  \\
        & $w_{gs}=4$ & 8.23               & 8.18                   & 8.24                 & 8.17                 & 8.37                  & \textbf{7.84}  \\
        & $w_{gs}=5$ & 7.97               & 8.03                   & 7.99                 & 7.92                 & 8.19                  & \textbf{7.74}  \\
        \midrule
        \multirow{5}{*}{Both} & Unguided   & 9.46               & 9.29                   & 9.50                 & 10.28                & 9.63                  & \textbf{9.15}  \\
        & $w_{gs}=2$ & 8.54               & 8.46                   & 8.60                 & 9.01                 & 8.56                  & \textbf{8.12}  \\
        & $w_{gs}=3$ & 8.19               & 8.13                   & 8.29                 & 8.43                 & 8.28                  & \textbf{7.96}  \\
        & $w_{gs}=4$ & 8.03               & 7.88                   & 8.25                 & 8.51                 & 8.08                  & \textbf{7.68}  \\
        & $w_{gs}=5$ & 8.01               & 7.85                   & 8.18                 & 8.48                 & 8.01                  & \textbf{7.71}  \\
        \bottomrule
    \end{tabular}
\end{table}

\subsection{Qualitative Evaluation}
\label{subsec:experiments-qualitative}
In addition to the quantitative evaluation based on a HTR downstream model, we show examples for generated images providing a qualitative evaluation in this section.
First, we present images sampled with different guidance scales in Tab.~\ref{tab:experiments-examples-gs}.
The images show known words from writers seen during the training on IAM\@.
The first, fourth and fifth row contain examples where unguided sampling resulted in erroneous spellings.
Increasing the guidance scale to $w_{gs}=3$ (row 4 \& 5) and $w_{gs}=5$ (row 1) resulted in correctly spelled words.
However in some cases (row 2), increasing the guidance scale did not improve the generation result.
In certain cases, guided sampling even led to worse spelling like shown in the example in row 3.
\begin{table}[tb]
    \caption{Examples for the influence of the guidance scale $w_{gs}$ for sampling on the generated images.
    The original image from the respective writer is shown in the first column as a reference.}
    \label{tab:experiments-examples-gs}
    \centering
    \begin{tabular}{C{2.5cm}|C{2.5cm}|C{2.5cm}|C{2.5cm}}
        \toprule
        Reference & Unguided & $w_{gs}=3$ & $w_{gs}=5$ \\
        \midrule
        \includegraphics[width=2.3cm]{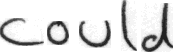} &
        \includegraphics[width=2.3cm]{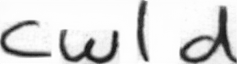} &
        \includegraphics[width=2.3cm]{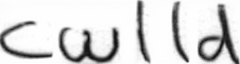} &
        \includegraphics[width=2.3cm]{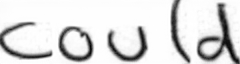} \\
        \includegraphics[width=2.3cm]{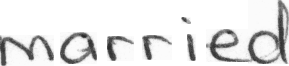} &
        \includegraphics[width=2.3cm]{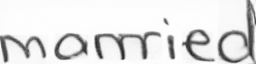} &
        \includegraphics[width=2.3cm]{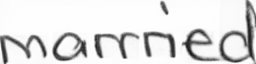} &
        \includegraphics[width=2.3cm]{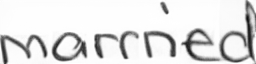} \\
        \includegraphics[width=2.3cm]{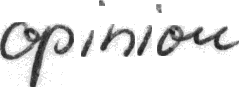} &
        \includegraphics[width=2.3cm]{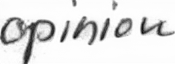} &
        \includegraphics[width=2.3cm]{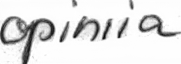} &
        \includegraphics[width=2.3cm]{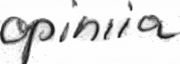} \\
        \includegraphics[width=2.3cm]{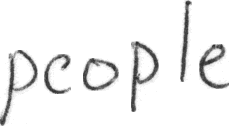} &
        \includegraphics[width=2.3cm]{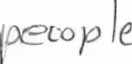} &
        \includegraphics[width=2.3cm]{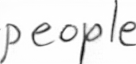} &
        \includegraphics[width=2.3cm]{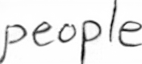} \\
        \includegraphics[width=2.3cm]{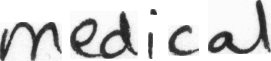} &
        \includegraphics[width=2.3cm]{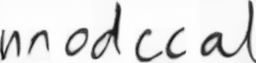} &
        \includegraphics[width=2.3cm]{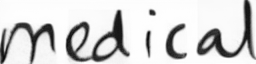} &
        \includegraphics[width=2.3cm]{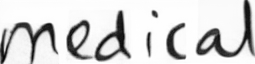} \\
        \bottomrule
    \end{tabular}
\end{table}

Using writer embeddings from a model instead of writer embeddings learned for specific writers allows our model to generalize to the the generation of writers not seen during training.
In the following we look at examples for the generation of unseen words (OOV) of writer seen during training (SW), known words (IV) of unseen writers (UW) and the combination (OOV-UW).
\begin{table}[tb]
    \caption{Examples for the generation of unseen words (OOV), unseen writers (UW) and the combination of both (OOV-UW). In-vocabulary words are denoted by IV and writers seen during training by SW.}
    \label{tab:experiments-examples-semi-supervised}
    \centering
    \begin{tabular}{L{2.5cm}|C{2.5cm}|C{2.5cm}|C{2.5cm}} %
        \toprule
        DM DS & IV-UW & OOV-SW & OOV-UW \\
        \midrule
        \multirow{2}{*}{IAM} &
        \includegraphics[width=2.3cm]{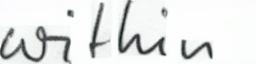} &
        \includegraphics[width=2.3cm]{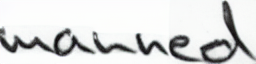} &
        \includegraphics[width=2.3cm]{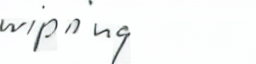} \\
        &
        \includegraphics[width=2.3cm]{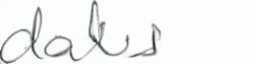} &
        \includegraphics[width=2.3cm]{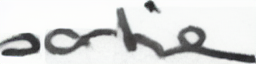} &
        \includegraphics[width=2.3cm]{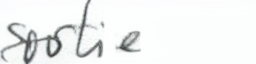} \\
        \midrule
        \multirow{4}{*}{\shortstack{IAM \& RIMES\\(semi-supervised)}} &
        \includegraphics[width=2.3cm]{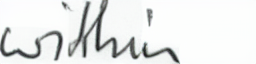} &
        \includegraphics[width=2.3cm]{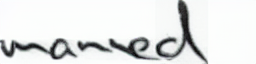} &
        \includegraphics[width=2.3cm]{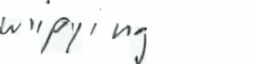} \\
        &
        \includegraphics[width=2.3cm]{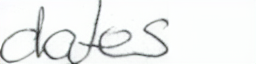} &
        \includegraphics[width=2.3cm]{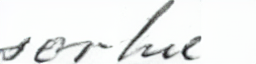} &
        \includegraphics[width=2.3cm]{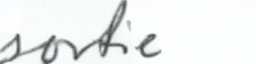} \\
        &
        \includegraphics[width=2.3cm]{dates_01552_Rimes_test_run_928} &
        \includegraphics[width=2.3cm]{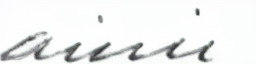} &
        \includegraphics[width=2.3cm]{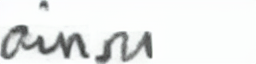} \\
        & &
        \includegraphics[width=2.3cm]{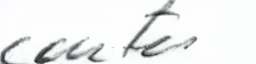} &
        \includegraphics[width=2.3cm]{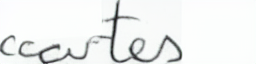} \\
        \midrule
        \multirow{3}{*}{\shortstack{IAM \& RIMES\\(fully-supervised)}} &
        \includegraphics[width=2.3cm]{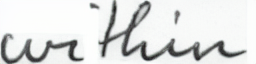} &
        \includegraphics[width=2.3cm]{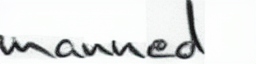} &
        \includegraphics[width=2.3cm]{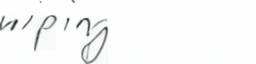} \\
        &
        \includegraphics[width=2.3cm]{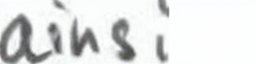} &
        \includegraphics[width=2.3cm]{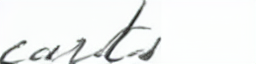} &
        \includegraphics[width=2.3cm]{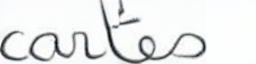} \\
    \end{tabular}
\end{table}
The first row of Tab.~\ref{tab:experiments-examples-semi-supervised} shows images with words and writers from IAM generated by model trained only on IAM\@.
While the IV-UW and the OOV-SW samples exhibit high quality, the spelling of \enquote{wiping} is erroneous.
The second row exemplary illustrates the generalization capabilities of the model to words\footnote{\enquote{dates} is part of the lexicon of RIMES and IAM.} and writers\footnote{Row 2 column 1 (OOV-SW) is a writer from IAM due to definition of the task.} of RIMES\@.
Especially for unseen writers the generated handwriting is messy and hard to read.
For the model trained in a semi-supervised fashion, row 3 depicts generated images for the same writers and words from IAM as row 1.
Rows 4 and 5 depict generated examples for writers from RIMES\@.
Examples for the words \enquote{ainsi} and \enquote{sortie} were seen during training, however without a label.
While the handwriting in row 4 is slightly better readable than the OOV-UW (writer and word from RIMES) example generated by the DM trained on IAM, row 5 reveals that the improvement does not hold for all cases.
Row 6 shows an example for the word \enquote{cartes} from RIMES which was not seen during training (not even without transcription), being not recognizable when combined with an unknown writer.
The last two rows show images generated by a DM that was trained with full supervision on IAM and RIMES for reference.
While outperforming both other models in terms of CER-train on RIMES (cf.~\ref{tab:experiments-RIMES-only-cer-train}), the visual quality of the examples for IAM (row 7) and RIMES (row 8) does not exceed that of the other models.
Overall, this qualitative assessment reveals that the visual quality of the generated images does not necessarily correspond to their suitability as training data for a HTR model.

\subsection{Comparison to SotA Approaches}
\label{subsec:experiments-related-work}
Finally, we compare our proposed approach to state-of-the art methods from the literature.
Wordstylist~\cite{Nikolaidou2023_WSV} which has been the basis for this work serves as a baseline.
Furthermore, we compare our method to DiffusionPen~\cite{Nikolaidou2024_DTC} where a similar modification regarding the computation of writer embeddings was proposed.
We use the same scenarios as in the previous experiments for evaluation.
First, we train the DM and the respective style encoder of the respective approach only on IAM and use CER-train to evaluate the generation performance for IAM and RIMES\@.
Additionally, we consider the scenario, where the fully labeled IAM database and the RIMES dataset without transcriptions is available.
The results are presented in Tab.~\ref{tab:experiments-related-work}.

\begin{table}[tb]
    \centering
    \caption{Comparision of our proposed method to Wordstylist~\cite{Nikolaidou2023_WSV} and DiffusionPen~\cite{Nikolaidou2024_DTC}.
    \enquote{DiffusionPen~\cite{Nikolaidou2024_DTC} (both)} denotes the experiment where their style encoder was trained on both datasets.}
    \label{tab:experiments-related-work}
    \begin{tabular}{L{12em}|C{6em}|C{3em}|R{4em}|R{4em}}
        \toprule
        Model & Noise Steps & $w_{gs}$ & \multicolumn{2}{c}{CER-train} \\
        &                                   &     & IAM           & RIMES         \\
        \midrule
        Wordstylist~\cite{Nikolaidou2023_WSV}         & 600~\cite{Ho2020_DDP}             & -   & 7.41          & -             \\
        Wordstylist~\cite{Nikolaidou2023_WSV}         & \phantom{0}50~\cite{Zhao2023_UUP} & -   & \textbf{7.36} & -     \\
        \midrule
        DiffusionPen~\cite{Nikolaidou2024_DTC}        & \phantom{0}50~\cite{Song2021_DDI} & -   & 7.70          & 24.16         \\
        DiffusionPen~\cite{Nikolaidou2024_DTC} (both) & \phantom{0}50~\cite{Song2021_DDI} & -   & 7.77 & 16.55 \\
        \midrule
        Ours (IAM only)                               & \phantom{0}50~\cite{Zhao2023_UUP} & 2.0 & 7.84          & 10.12         \\
        Ours (IAM only)                               & \phantom{0}50~\cite{Zhao2023_UUP} & 5.0 & 7.39          & 9.99          \\
        Ours (semi-supervised)                        & \phantom{0}50~\cite{Zhao2023_UUP} & 2.0 & 8.15          & \textbf{6.61} \\
        Ours (semi-supervised)                        & \phantom{0}50~\cite{Zhao2023_UUP} & 5.0 & 7.74          & 6.91          \\
        \bottomrule
    \end{tabular}
\end{table}

For our approach, we trained the diffusion model using classifier-free guidance with $p_\text{uncond}=0.1$ and \addtotimestep{} for incorporating the style embedding.
In~\cite{Nikolaidou2023_WSV}, sampling was performed 600 steps in the reverse process.
We additionally used the weights provided with the published reference implementation\footnote{\url{https://github.com/koninik/WordStylist}} to evaluate the model when using UniPC~\cite{Zhao2023_UUP} with 50 noise steps like in our experimental setup.
In Tab.~\ref{tab:experiments-related-work}, we can see that the influence of the reduction of sampling steps is only minimal.
Due to the limitations to writers seen during training we can only compare to Wordstylist on IAM where our approach performs comparable when trained only on IAM and slightly worse when trained semi-supervised on IAM and RIMES\@.

Since DiffusionPen~\cite{Nikolaidou2024_DTC} was trained on a different subset of IAM\footnote{Although the RWTH Aachen split was used, the subset of IAM used for training is slightly different.
Addtionally, longer words were considered and more characters were included},
we trained the model based on the published reference implementation\footnote{\url{https://github.com/koninik/DiffusionPen}} but using our data partitioning.
On IAM, DiffusionPen performs comparable to Wordstylist and our approach.
Since the introduced style encoder enables the approach to generate examples of unseen writers, we used the model trained on IAM to generate training data of RIMES\@.
However, the model performed poorly in this task and is clearly outperformed by our approach.
For a more fair comparison, we conducted an additional experiment where we trained their style encoder on both datasets as we assume to have writer labels available for RIMES\@.
While there is almost no effect on the generation performance of IAM, the model generalizes much better to the generation of examples from the RIMES data set.
Nevertheless, especially our semi-supervised approach achieves considerably better results in this scenario.

    \section{Conclusion}
\label{sec:conclusion}
In this work, we have presented a method for handwritten text generation based on diffusion models.
By computing style embeddings with a masked autoencoder trained on handwriting images, we have enabled our DM to generate word images of writers not seen during training.
Additionally, we explored different ways of incorporating the style and the text conditioning into the generation process and employed classifier-free guidance for better generation results.
For the interesting application of generation of training data for new datasets where transcriptions are unknown, we have proposed a semi-supervised training scheme for our DM resulting in improved generation quality for the new dataset.
In our evaluation, we however showed that while being suitable as training data for a handwriting recognition model, the visual quality of the generated images is still poor in some cases.
Furthermore, there is still a gap in performance to using training data generated by a model which was trained with full supervision.
Therefore, further research on adaptation of HTG models for the generation of new and unlabeled datasets is of high interest especially due to one motivation of HTG being the generation of training material.

    \bibliographystyle{splncs04}
    \bibliography{literature}

\begin{thebibliography}{10}
\providecommand{\url}[1]{\texttt{#1}}
\providecommand{\urlprefix}{URL }
\providecommand{\doi}[1]{https://doi.org/#1}

\bibitem{Aksan2018_DMD}
Aksan, E., Pece, F., Hilliges, O.: {DeepWriting: Making Digital Ink Editable
  via Deep Generative Modeling}. In: Proc. CHI Conference on Human Factors in
  Computing Systems. Montreal, Canada (2018). \doi{10.1145/3173574.3173779}

\bibitem{Alonso2019_AGH}
Alonso, E., Moysset, B., Messina, R.: {Adversarial Generation of Handwritten
  Text Images Conditioned on Sequences}. In: Int. Conf. on Document Analysis
  and Recognition. pp. 481--486. Sydney, Australia (2019).
  \doi{10.1109/ICDAR.2019.00083}

\bibitem{Ba2016_LN}
Ba, J.L., Kiros, J.R., Hinton, G.E.: {Layer Normalization}. arXiv:
  abs/1607.06450  (2016). \doi{10.48550/arXiv.1607.06450}

\bibitem{Bhunia2021_HT}
Bhunia, A., Khan, S., Cholakkal, H., Anwer, R., Khan, F., Shah, M.:
  {Handwriting Transformers}. In: IEEE/CVF Int. Conf. on Computer Vision. pp.
  1066--1074. Los Alamitos, CA, USA (2021). \doi{10.1109/ICCV48922.2021.00112}

\bibitem{Bishop2024_DLF}
Bishop, C.M., Bishop, H.: {Deep Learning: Foundations and Concepts}. Springer
  International Publishing (2024). \doi{10.1007/978-3-031-45468-4}

\bibitem{Chang2023_CHT}
Chang, C.C., Perera, L.P.G., Khudanpur, S.: {Crosslingual Handwritten Text
  Generation Using GANs}. In: Int. Conf. on Document Analysis and Recognition
  Workshops. pp. 285--301. San José, CA, USA (2023).
  \doi{10.1007/978-3-031-41501-2\_20}

\bibitem{Chung2015_RLV}
Chung, J., Kastner, K., Dinh, L., Goel, K., Courville, A., Bengio, Y.: {A
  Recurrent Latent Variable Model for Sequential Data}. In: Advances in Neural
  Information Processing Systems. Montréal, Canada (2015)

\bibitem{Dai2024_OSD}
Dai, G., Zhang, Y., Ke, Q., Guo, Q., Huang, S.: {One-Shot Diffusion Mimicker
  for Handwritten Text Generation}. In: European Conf. on Computer Vision. pp.
  410--427. Milan, Italy (2024). \doi{10.1007/978-3-031-73636-0\_24}

\bibitem{Davis2020_TSC}
Davis, B.L., Tensmeyer, C., Price, B.L., Wigington, C., Morse, B.S., Jain, R.:
  {Text and Style Conditioned GAN for Generation of Offline Handwriting Lines}.
  In: British Machine Vision Conference. Virtual Conference (2020)

\bibitem{Deng2009_Ils}
Deng, J., Dong, W., Socher, R., Li, L.J., Li, K., Fei-Fei, L.: {ImageNet: A
  large-scale hierarchical image database}. In: IEEE/CVF Conf. on Computer
  Vision and Pattern Recognition. pp. 248--255. Miami, FL, USA (2009).
  \doi{10.1109/CVPR.2009.5206848}

\bibitem{Dhariwal2021_DMB}
Dhariwal, P., Nichol, A.: {Diffusion Models Beat GANs on Image Synthesis}. In:
  Advances in Neural Information Processing Systems. pp. 8780--8794. Virtual
  Conference (2021)

\bibitem{Ding2023_IHO}
Ding, H., Luan, B., Gui, D., Chen, K., Huo, Q.: {Improving Handwritten OCR with
  Training Samples Generated by Glyph Conditional Denoising Diffusion
  Probabilistic Model}. In: Int. Conf. on Document Analysis and Recognition.
  pp. 20--37. San José, CA, USA (2023). \doi{10.1007/978-3-031-41685-9\_2}

\bibitem{Dosovitskiy2021_IiW}
Dosovitskiy, A., Beyer, L., Kolesnikov, A., Weissenborn, D., Zhai, X.,
  Unterthiner, T., Dehghani, M., Minderer, M., Heigold, G., Gelly, S.,
  Uszkoreit, J., Houlsby, N.: {An Image is Worth 16x16 Words: Transformers for
  Image Recognition at Scale}. In: Int. Conf. on Learning Representations.
  Vienna, Austria (2021)

\bibitem{Elanwar2024_Gan}
Elanwar, R., Betke, M.: {Generative Adversarial Networks for Handwriting Image
  Generation: A Review}. The Visual Computer  (2024).
  \doi{10.1007/s00371-024-03534-9}

\bibitem{Esser2021_TTH}
Esser, P., Rombach, R., Ommer, B.: {Taming Transformers for High-Resolution
  Image Synthesis}. In: IEEE/CVF Conf. on Computer Vision and Pattern
  Recognition. pp. 12868--12878. Virtual Conference (2021).
  \doi{10.1109/CVPR46437.2021.01268}

\bibitem{Fogel2020_SSS}
Fogel, S., Averbuch-Elor, H., Cohen, S., Mazor, S., Litman, R.: {ScrabbleGAN:
  Semi-Supervised Varying Length Handwritten Text Generation}. In: IEEE/CVF
  Conf. on Computer Vision and Pattern Recognition. pp. 4323--4332. Seattle,
  WA, USA (2020). \doi{10.1109/CVPR42600.2020.00438}

\bibitem{Gan2021_HHI}
Gan, J., Wang, W.: {HiGAN: Handwriting Imitation Conditioned on
  Arbitrary-Length Texts and Disentangled Styles}. In: Proc. AAAI Conf. on
  Artificial Intelligence. pp. 7484--7492. Virtual Conference (2021).
  \doi{10.1609/aaai.v35i9.16917}

\bibitem{Gan2022_HHI}
Gan, J., Wang, W., Leng, J., Gao, X.: {HiGAN+: Handwriting Imitation GAN with
  Disentangled Representations}. ACM Trans. Graph.  \textbf{42}(1),  1--17
  (2022). \doi{10.1145/3550070}

\bibitem{Ganin2018_SPI}
Ganin, Y., Kulkarni, T., Babuschkin, I., Eslami, S.M.A., Vinyals, O.:
  {Synthesizing Programs for Images using Reinforced Adversarial Learning}. In:
  Int. Conf. on Machine Learning. pp. 1666--1675. Stockholm, Sweden (2018)

\bibitem{Goodfellow2014_GAN}
Goodfellow, I., Pouget-Abadie, J., Mirza, M., Xu, B., Warde-Farley, D., Ozair,
  S., Courville, A., Bengio, Y.: {Generative Adversarial Nets}. In: Advances in
  Neural Information Processing Systems. Montréal, Canada (2014)

\bibitem{Graves2013_GSR}
Graves, A.: {Generating Sequences With Recurrent Neural Networks}. arXiv:
  abs/1308.0850  (2013). \doi{10.48550/arXiv.1308.0850}

\bibitem{Graves2006_Ctc}
Graves, A., Fern\'{a}ndez, S., Gomez, F., Schmidhuber, J.: {Connectionist
  Temporal Classification: Labelling Unsegmented Sequence Data with Recurrent
  Neural Networks}. In: Int. Conf. on Machine Learning. pp. 369--376.
  Pittsburgh, PA, USA (2006). \doi{10.1145/1143844.1143891}

\bibitem{Grosicki2009_I2H}
Grosicki, E., Abed, H.E.: {ICDAR 2009 Handwriting Recognition Competition}. In:
  Int. Conf. on Document Analysis and Recognition. pp. 1398--1402. Barcelona,
  Spain (2009). \doi{10.1109/ICDAR.2009.184}

\bibitem{Guan2020_IHO}
Guan, M., Ding, H., Chen, K., Huo, Q.: {Improving Handwritten OCR with
  Augmented Text Line Images Synthesized from Online Handwriting Samples by
  Style-Conditioned GAN}. In: Int. Conf. on Frontiers in Handwriting
  Recognition. pp. 151--156. Virtual Conference (2020).
  \doi{10.1109/ICFHR2020.2020.00037}

\bibitem{Gui2023_ZsG}
Gui, D., Chen, K., Ding, H., Huo, Q.: {Zero-shot Generation of Training Data
  with Denoising Diffusion Probabilistic Model for Handwritten Chinese
  Character Recognition}. In: Int. Conf. on Document Analysis and Recognition.
  pp. 348--365. San José, CA, USA (2023). \doi{10.1007/978-3-031-41679-8\_20}

\bibitem{Haines2016_MTY}
Haines, T.S.F., Mac~Aodha, O., Brostow, G.J.: {My Text in Your Handwriting}.
  ACM Trans. on Graphics  \textbf{35}(3),  1--18 (2016). \doi{10.1145/2886099}

\bibitem{He2022_MAA}
He, K., Chen, X., Xie, S., Li, Y., Doll\'ar, P., Girshick, R.: {Masked
  Autoencoders Are Scalable Vision Learners}. In: IEEE/CVF Conf. on Computer
  Vision and Pattern Recognition. pp. 15979--15988. New Orleans, LA, USA
  (2022). \doi{10.1109/CVPR52688.2022.01553}

\bibitem{He2016_DRL}
He, K., Zhang, X., Ren, S., Sun, J.: {Deep Residual Learning for Image
  Recognition}. In: IEEE/CVF Conf. on Computer Vision and Pattern Recognition.
  pp. 770--778. Las Vegas, NV, USA (2016)

\bibitem{Heusel2017_GTT}
Heusel, M., Ramsauer, H., Unterthiner, T., Nessler, B., Hochreiter, S.: {GANs
  Trained by a Two Time-Scale Update Rule Converge to a Local Nash
  Equilibrium}. In: Advances in Neural Information Processing Systems. Long
  Beach, CA, USA (2017)

\bibitem{Ho2020_DDP}
Ho, J., Jain, A., Abbeel, P.: {Denoising Diffusion Probabilistic Models}. In:
  Advances in Neural Information Processing Systems. Virtual Conference (2020)

\bibitem{Ho2022_CFD}
Ho, J., Salimans, T.: {Classifier-Free Diffusion Guidance}. In: NeurIPS
  Workshop on Deep Generative Models and Downstream Applications (2021).
  \doi{10.48550/arXiv.2207.12598}

\bibitem{Hochreiter1997_LST}
Hochreiter, S., Schmidhuber, J.: {Long Short-Term Memory}. Neural Computing
  \textbf{9}(8),  1735--1780 (1997). \doi{10.1162/neco.1997.9.8.1735}

\bibitem{Ingle2019_SHT}
Ingle, R.R., Fujii, Y., Deselaers, T., Baccash, J., Popat, A.C.: {A Scalable
  Handwritten Text Recognition System}. In: Int. Conf. on Document Analysis and
  Recognition. pp. 17--24. Sydney, Australia (2019).
  \doi{10.1109/ICDAR.2019.00013}

\bibitem{Isola2017_IIT}
Isola, P., Zhu, J.Y., Zhou, T., Efros, A.A.: {Image-To-Image Translation With
  Conditional Adversarial Networks}. In: IEEE/CVF Conf. on Computer Vision and
  Pattern Recognition. pp. 5967--5976. Honolulu, HI, USA (2017).
  \doi{10.1109/CVPR.2017.632}

\bibitem{Ji2019_GAN}
Ji, B., Chen, T.: {Generative Adversarial Network for Handwritten Text}. arXiv:
  abs/1907.11845  (2019). \doi{10.48550/arXiv.1907.11845}

\bibitem{Kang2020_DCS}
Kang, L., Riba, P., Rusiñol, M., Fornés, A., Villegas, M.: {Distilling
  Content from Style for Handwritten Word Recognition}. In: Int. Conf. on
  Frontiers in Handwriting Recognition. pp. 139--144. Virtual Conference
  (2020). \doi{10.1109/ICFHR2020.2020.00035}

\bibitem{Kang2022_CSA}
Kang, L., Riba, P., Rusiñol, M., Fornés, A., Villegas, M.: {Content and Style
  Aware Generation of Text-Line Images for Handwriting Recognition}. IEEE
  Trans. on Pattern Analysis and Machine Intelligence  \textbf{44}(12),
  8846--8860 (2022). \doi{10.1109/TPAMI.2021.3122572}

\bibitem{Kang2020_GCC}
Kang, L., Riba, P., Wang, Y., Rusi\~{n}ol, M., Forn\'{e}s, A., Villegas, M.:
  {GANwriting: Content-Conditioned Generation of Styled Handwritten Word
  Images}. In: European Conf. on Computer Vision. pp. 273--289. Glasgow, United
  Kingdom (2020). \doi{10.1007/978-3-030-58592-1\_17}

\bibitem{Kang2020_UWA}
Kang, L., Rusinol, M., Fornes, A., Riba, P., Villegas, M.: {Unsupervised Writer
  Adaptation for Synthetic-to-Real Handwritten Word Recognition}. In: Proc. of
  the IEEE/CVF Winter Conf. on Applications of Computer Vision. pp. 3491--3500.
  Snowmass Village, CO, USA (2020). \doi{10.1109/WACV45572.2020.9093392}

\bibitem{Kang2019_CAS}
Kang, L., Toledo, J.I., Riba, P., Villegas, M., Fornés, A., Rusiñol, M.:
  {Convolve, Attend and Spell: An Attention-based Sequence-to-Sequence Model
  for Handwritten Word Recognition}. In: Proc. German Conf. on Pattern
  Recognition. pp. 459--472. Stuttgart, Germany (2018).
  \doi{10.1007/978-3-030-12939-2_32}

\bibitem{Konidaris2007_Kgw}
Konidaris, T., Gatos, B., Ntzios, K., Pratikakis, I., Theodoridis, S.,
  Perantonis, S.J.: {Keyword-Guided Word Spotting in Historical Printed
  Documents Using Synthetic Data and User Feedback}. Int. Journal on Document
  Analysis and Recognition  \textbf{9}(2–4),  167--177 (Mar 2007).
  \doi{10.1007/s10032-007-0042-4}

\bibitem{Krishnan2016_GSD}
Krishnan, P., Jawahar, C.V.: {Generating Synthetic Data for Text Recognition}.
  arXiv: abs/1608.04224  (2016). \doi{10.48550/arXiv.1608.04224}

\bibitem{Krishnan2019_Hve}
Krishnan, P., Jawahar, C.V.: {HWNet v2: an efficient word image representation
  for handwritten documents}. Int. Journal on Document Analysis and Recognition
   \textbf{22}(4),  387–405 (2019). \doi{10.1007/s10032-019-00336-x}

\bibitem{Krishnan2021_TTT}
Krishnan, P., Kovvuri, R., Pang, G., Vassilev, B., Hassner, T.:
  {TextStyleBrush: Transfer of Text Aesthetics from a Single Example}. arXiv:
  abs/2106.08385  (2021). \doi{10.48550/ARXIV.2106.08385}, 2023 pusblished in
  PAMI

\bibitem{Lee2019_STF}
Lee, J., Lee, Y., Kim, J., Kosiorek, A., Choi, S., Teh, Y.W.: {Set Transformer:
  A Framework for Attention-based Permutation-Invariant Neural Networks}. In:
  Int. Conf. on Machine Learning. pp. 3744--3753. Long Beach, CA, USA (2019)

\bibitem{Li2021_TTb}
Li, M., Lv, T., Chen, J., Cui, L., Lu, Y., Florencio, D., Zhang, C., Li, Z.,
  Wei, F.: {TrOCR: Transformer-based Optical Character Recognition with
  Pre-trained Models}. In: Proc. AAAI Conf. on Artificial Intelligence. pp.
  13094--13102. Washington DC, USA (2023)

\bibitem{Lin2007_SpE}
Lin, Z., Wan, L.: {Style-preserving English handwriting synthesis}. Pattern
  Recognition  \textbf{40}(7),  2097--2109 (2007).
  \doi{10.1016/j.patcog.2006.11.024}

\bibitem{Liu2021_HTG}
Liu, X., Meng, G., Xiang, S., Pan, C.: {Handwritten Text Generation via
  Disentangled Representations}. IEEE Signal Processing Letters  \textbf{28},
  1838--1842 (2021). \doi{10.1109/LSP.2021.3109541}

\bibitem{Loshchilov2019_DWD}
Loshchilov, I., Hutter, F.: {Decoupled Weight Decay Regularization}. In: Int.
  Conf. on Learning Representations. New Orleans, LA, USA (2019)

\bibitem{Lu2022_DSF}
Lu, C., Zhou, Y., Bao, F., Chen, J., Chongxuan, L.I., Zhu, J.: {DPM-Solver: A
  Fast ODE Solver for Diffusion Probabilistic Model Sampling in Around 10
  Steps}. In: Advances in Neural Information Processing Systems. pp.
  5775--5787. New Orleans, LA, USA (2022)

\bibitem{Lu2022_DSFa}
Lu, C., Zhou, Y., Bao, F., Chen, J., Li, C., Zhu, J.: {DPM-Solver++: Fast
  Solver for Guided Sampling of Diffusion Probabilistic Models}. arXiv:
  abs/2211.01095  (2022). \doi{10.48550/arXiv.2211.01095}

\bibitem{Luhman2020_DmH}
Luhman, T., Luhman, E.: {Diffusion models for Handwriting Generation}. arXiv:
  abs/2011.06704  (2020). \doi{10.48550/arXiv.2011.06704}

\bibitem{Luo2023_SHS}
Luo, C., Zhu, Y., Jin, L., Li, Z., Peng, D.: {SLOGAN: Handwriting Style
  Synthesis for Arbitrary-Length and Out-of-Vocabulary Text}. IEEE Trans. on
  Neural Networks and Learning Systems  \textbf{34}(11),  8503--8515 (2023).
  \doi{10.1109/TNNLS.2022.3151477}

\bibitem{Marti2002_IdE}
Marti, U.V., Bunke, H.: {The IAM-database: an English sentence database for
  offline handwriting recognition}. Int. Journal on Document Analysis and
  Recognition  \textbf{5}(1),  39--46 (2002). \doi{10.1007/s100320200071}

\bibitem{Mattick2021_SIH}
Mattick, A., Mayr, M., Seuret, M., Maier, A., Christlein, V.: {SmartPatch:
  Improving Handwritten Word Imitation with Patch Discriminators}. In:
  Llad{\'o}s, J., Lopresti, D., Uchida, S. (eds.) Int. Conf. on Document
  Analysis and Recognition. pp. 268--283. Lausanne, Switzerland (2021).
  \doi{10.1007/978-3-030-86549-8\_18}

\bibitem{Mayr2024_ZSP}
Mayr, M., Dreier, M., Kordon, F., Seuret, M., Zöllner, J., Wu, F., Maier, A.,
  Christlein, V.: {Zero-Shot Paragraph-level Handwriting Imitation with Latent
  Diffusion Models}. arXiv: abs/2409.00786  (2024).
  \doi{10.48550/arXiv.2409.00786}

\bibitem{Mayr2020_STH}
Mayr, M., Stumpf, M., Nicolaou, A., Seuret, M., Maier, A., Christlein, V.:
  {Spatio-Temporal Handwriting Imitation}. In: European Conf. on Computer
  Vision Workshops. pp. 528--543. Glasgow, United Kingdom (2020).
  \doi{10.1007/978-3-030-68238-5_38}

\bibitem{Michael2019_ESS}
Michael, J., Labahn, R., Grüning, T., Zöllner, J.: {Evaluating
  Sequence-to-Sequence Models for Handwritten Text Recognition}. In: Int. Conf.
  on Document Analysis and Recognition. pp. 1286--1293. Sydney, Australia
  (2019). \doi{10.1109/ICDAR.2019.00208}

\bibitem{Mirza2014_CGA}
Mirza, M., Osindero, S.: {Conditional Generative Adversarial Nets}. arXiv:
  abs/1411.1784  (2014). \doi{10.48550/arXiv.1411.1784}

\bibitem{Miyato2018_cPD}
Miyato, T., Koyama, M.: {cGANs with Projection Discriminator}. In: Int. Conf.
  on Learning Representations. Vancouver, Canada (2018)

\bibitem{Nichol2021_IDD}
Nichol, A.Q., Dhariwal, P.: {Improved Denoising Diffusion Probabilistic Model}.
  In: Int. Conf. on Machine Learning. pp. 8162--8171. Virtual Conference (2021)

\bibitem{Nichol2022_GTP}
Nichol, A.Q., Dhariwal, P., Ramesh, A., Shyam, P., Mishkin, P., Mcgrew, B.,
  Sutskever, I., Chen, M.: {GLIDE: Towards Photorealistic Image Generation and
  Editing with Text-Guided Diffusion Models}. In: Int. Conf. on Machine
  Learning. pp. 16784--16804. Baltimore, MD, USA (2022)

\bibitem{Nikolaidou2023_WSV}
Nikolaidou, K., Retsinas, G., Christlein, V., Seuret, M., Sfikas, G., Smith,
  E.B., Mokayed, H., Liwicki, M.: {Wordstylist: Styled Verbatim Handwritten
  Text Generation with Latent Diffusion Models}. In: Int. Conf. on Document
  Analysis and Recognition. pp. 384--401. San José, CA, USA (2023).
  \doi{10.1007/978-3-031-41679-8\_22}

\bibitem{Nikolaidou2024_DTC}
Nikolaidou, K., Retsinas, G., Sfikas, G., Liwicki, M.: {DiffusionPen: Towards
  Controlling the Style of Handwritten Text Generation}. In: European Conf. on
  Computer Vision. pp. 417--434. Milan, Italy (2024).
  \doi{10.1007/978-3-031-73013-9\_24}

\bibitem{Oord2017_NDR}
van~den Oord, A., Vinyals, O., kavukcuoglu, k.: {Neural Discrete Representation
  Learning}. In: Advances in Neural Information Processing Systems. Long Beach,
  CA, USA (2017)

\bibitem{Pippi2023_HTG}
Pippi, V., Cascianelli, S., Cucchiara, R.: {Handwritten Text Generation From
  Visual Archetypes}. In: IEEE/CVF Conf. on Computer Vision and Pattern
  Recognition. pp. 22458--22467. Vancouver, Canada (2023)

\bibitem{Retsinas2022_BPH}
Retsinas, G., Sfikas, G., Gatos, B., Nikou, C.: {Best Practices for a
  Handwritten Text Recognition System}. In: Int. Workshop on Document Analysis
  Systems. pp. 247--259. La Rochelle, France (2022).
  \doi{10.1007/978-3-031-06555-2\_17}

\bibitem{Riaz2024_SSA}
Riaz, N., Saifullah, S., Agne, S., Dengel, A., Ahmed, S.: {StylusAI: Stylistic
  Adaptation for Robust German Handwritten Text Generation}. In: Int. Conf. on
  Document Analysis and Recognition. pp. 429--444. Athens, Greece (2024).
  \doi{10.1007/978-3-031-70536-6\_26}

\bibitem{Rombach2022_HRI}
Rombach, R., Blattmann, A., Lorenz, D., Esser, P., Ommer, B.: {High-Resolution
  Image Synthesis With Latent Diffusion Models}. In: IEEE/CVF Conf. on Computer
  Vision and Pattern Recognition. pp. 10684--10695. New Orleans, LA, USA (2022)

\bibitem{Ronneberger2015_UNC}
Ronneberger, O., Fischer, P., Brox, T.: {U-Net: Convolutional Networks for
  Biomedical Image Segmentation}. In: Int. Conf. on Medical Image Computing and
  Computer-Assisted Intervention. pp. 234--241. Munich, Germany (2015).
  \doi{10.1007/978-3-319-24574-4\_28}

\bibitem{Sandler2018_MIR}
Sandler, M., Howard, A., Zhu, M., Zhmoginov, A., Chen, L.C.: {MobileNetV2:
  Inverted Residuals and Linear Bottlenecks}. In: IEEE/CVF Conf. on Computer
  Vision and Pattern Recognition. pp. 4510--4520. Salt Lake City, UT, USA
  (2018). \doi{10.1109/CVPR.2018.00474}

\bibitem{SohlDickstein2015_DUL}
Sohl-Dickstein, J., Weiss, E., Maheswaranathan, N., Ganguli, S.: {Deep
  Unsupervised Learning using Nonequilibrium Thermodynamics}. In: Int. Conf. on
  Machine Learning. pp. 2256--2265. Lille, France (2015)

\bibitem{Song2021_DDI}
Song, J., Meng, C., Ermon, S.: {Denoising Diffusion Implicit Models}. In: Int.
  Conf. on Learning Representations. Vienna, Austria (2021)

\bibitem{Song2019_GME}
Song, Y., Ermon, S.: {Generative Modeling by Estimating Gradients of the Data
  Distribution}. In: Advances in Neural Information Processing Systems.
  Vancouver, Canada (2019)

\bibitem{Song2021_SBG}
Song, Y., Sohl-Dickstein, J., Kingma, D.P., Kumar, A., Ermon, S., Poole, B.:
  {Score-Based Generative Modeling through Stochastic Differential Equations}.
  In: Int. Conf. on Learning Representations. Vienna, Austria (2021)

\bibitem{Souibgui2023_TDs}
Souibgui, M.A., Biswas, S., Mafla, A., Biten, A.F., Forn\'{e}s, A., Kessentini,
  Y., Llad\'{o}s, J., Gomez, L., Karatzas, D.: {Text-DIAE: a self-supervised
  degradation invariant autoencoder for text recognition and document
  enhancement}. In: Proc. AAAI Conf. on Artificial Intelligence. pp.
  2330--2338. Washington DC, USA (2023). \doi{10.1609/aaai.v37i2.25328}

\bibitem{Szegedy2016_RIA}
Szegedy, C., Vanhoucke, V., Ioffe, S., Shlens, J., Wojna, Z.: {Rethinking the
  Inception Architecture for Computer Vision}. In: IEEE/CVF Conf. on Computer
  Vision and Pattern Recognition. pp. 2818--2826. Las Vegas, NV, USA (2016).
  \doi{10.1109/CVPR.2016.308}

\bibitem{Thomas2009_ShC}
Thomas, A.O., Rusu, A., Govindaraju, V.: {Synthetic handwritten CAPTCHAs}.
  Pattern Recognition  \textbf{42}(12),  3365--3373 (2009).
  \doi{10.1016/j.patcog.2008.12.018}

\bibitem{Vanherle2024_VCY}
Vanherle, B., Pippi, V., Cascianelli, S., Michiels, N., Van~Reeth, F.,
  Cucchiara, R.: {VATr++: Choose Your Words Wisely for Handwritten Text
  Generation}. arXiv: abs/2402.10798  (2024). \doi{10.48550/arXiv.2402.10798}

\bibitem{Vaswani2017_AiA}
Vaswani, A., Shazeer, N., Parmar, N., Uszkoreit, J., Jones, L., Gomez, A.N.,
  Kaiser, L., Polosukhin, I.: {Attention is All you Need}. In: Advances in
  Neural Information Processing Systems. Long Beach, CA, USA (2017)

\bibitem{Wang2005_Csp}
Wang, J., Wu, C., Xu, Y.Q., Shum, H.Y.: {Combining Shape and Physical Models
  for Online Cursive Handwriting Synthesis}. Int. Journal on Document Analysis
  and Recognition  \textbf{7}(4),  219--227 (2005).
  \doi{10.1007/s10032-004-0131-6}

\bibitem{Wang2022_ABT}
Wang, Y., Wang, H., Sun, S., Wei, H.: {An Approach Based on Transformer and
  Deformable Convolution for Realistic Handwriting Samples Generation}. In:
  Int. Conf. on Pattern Recognition. pp. 1457--1463. Montréal, Canada (2022).
  \doi{10.1109/ICPR56361.2022.9956551}

\bibitem{Zdenek2021_JME}
Zdenek, J., Nakayama, H.: {JokerGAN: Memory-Efficient Model for Handwritten
  Text Generation with Text Line Awareness}. In: ACM Int. Conf. on Multimedia.
  pp. 5655--5663. Virtual Conference (2021). \doi{10.1145/3474085.3475713}

\bibitem{Zdenek2023_HTG}
Zdenek, J., Nakayama, H.: {Handwritten Text Generation with Character-Specific
  Encoding for Style Imitation}. In: Int. Conf. on Document Analysis and
  Recognition. pp. 313--329. San José, CA, USA (2023).
  \doi{10.1007/978-3-031-41679-8\_18}

\bibitem{Zhang2023_FSD}
Zhang, Q., Chen, Y.: {Fast Sampling of Diffusion Models with Exponential
  Integrator}. In: Int. Conf. on Learning Representations. Kigali, Rwanda
  (2023)

\bibitem{Zhang2018_UED}
Zhang, R., Isola, P., Efros, A.A., Shechtman, E., Wang, O.: {The Unreasonable
  Effectiveness of Deep Features as a Perceptual Metric}. In: IEEE/CVF Conf. on
  Computer Vision and Pattern Recognition. pp. 586--595. Salt Lake City, UT,
  USA (2018). \doi{10.1109/CVPR.2018.00068}

\bibitem{Zhao2023_UUP}
Zhao, W., Bai, L., Rao, Y., Zhou, J., Lu, J.: {UniPC: A Unified
  Predictor-Corrector Framework for Fast Sampling of Diffusion Models}. In:
  Advances in Neural Information Processing Systems. New Orleans, LA, USA
  (2023)

\bibitem{Zhu2023_CTI}
Zhu, Y., Li, Z., Wang, T., He, M., Yao, C.: {Conditional Text Image Generation
  With Diffusion Models}. In: IEEE/CVF Conf. on Computer Vision and Pattern
  Recognition. pp. 14235--14245. Vancouver, Canada (2023)

\end{thebibliography}
\end{document}